%% file: main.tex
\documentclass[runningheads]{llncs}

\usepackage{epsfig}
\usepackage{comment}
\usepackage{amsmath}
\usepackage{amssymb}
\usepackage{bm}
\usepackage{xspace}
\usepackage{graphicx}  \graphicspath{{figures/}}
\usepackage{algorithm}
\usepackage{algpseudocode}
\usepackage[normalem]{ulem}
\usepackage{makecell}
\usepackage{multirow}
\usepackage{rotating}
\usepackage{booktabs}
\usepackage{capt-of}
\usepackage{enumitem}
\usepackage{microtype}
\usepackage[dvipsnames,svgnames,x11names]{xcolor}
\usepackage{subcaption}

\usepackage[pagebackref=true,breaklinks=true,letterpaper=true,colorlinks,bookmarks=false]{hyperref}

\input{macros}

\mainmatter
\def\ECCVSubNumber{4452}  

\begin{document}

\title{DeepGMR: Learning Latent Gaussian Mixture Models for Registration}

\titlerunning{DeepGMR: Learning Latent Gaussian Mixture Models for Registration} 
\authorrunning{Wentao Yuan, Ben Eckart, Kihwan Kim \etal} 
\author{Wentao Yuan\inst{1}* \quad  Ben Eckart\inst{2} \quad Kihwan Kim\inst{2}\\Varun Jampani\inst{2} \quad Dieter Fox\inst{1,2} \quad Jan Kautz\inst{2}}
\institute{University of Washington \and NVIDIA}

\maketitle

\begin{center}
\vspace{6mm}
\centerline{
\includegraphics[width=0.95\textwidth,trim={4pt 35pt 4pt 30pt}]{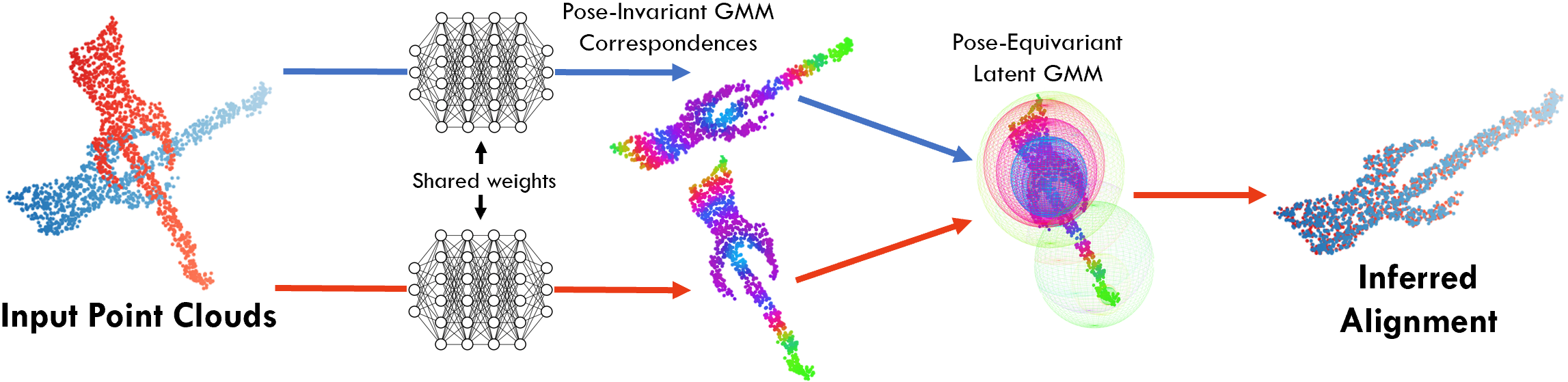}}
\captionof{figure}
{
 DeepGMR aligns point clouds with arbitrary poses by predicting pose-invariant point correspondences to a learned latent Gaussian Mixture Model (GMM) distribution.
}
\vspace{-10mm}
\label{fig:teaser}
\end{center}
\blfootnote{*Work partially done during an internship at NVIDIA}
\blfootnote{Code and data are available at \href{https://wentaoyuan.github.io/deepgmr}{https://wentaoyuan.github.io/deepgmr}}

\input{sec_abstract}

\input{sec_introduction}

\input{sec_related}

\input{sec_preliminaries}

\input{sec_approach}

\input{sec_results}

\input{sec_conclusion}

\clearpage

\bibliographystyle{splncs04}
\bibliography{references}

\title{DeepGMR: Learning Latent Gaussian Mixture Models for Registration\\--Supplementary Material--}
\titlerunning{DeepGMR: Learning Latent Gaussian Mixture Models for Registration} 
\authorrunning{Wentao Yuan, Ben Eckart, Kihwan Kim \etal} 
\author{Wentao Yuan\inst{1} \quad  Ben Eckart\inst{2} \quad Kihwan Kim\inst{2}\\Varun Jampani\inst{2} \quad Dieter Fox\inst{1,2} \quad Jan Kautz\inst{2}}
\institute{University of Washington \and NVIDIA}

\maketitle

\input{supp_overview.tex}

\input{supp_math.tex}

\input{supp_quant.tex}

\input{supp_qual.tex}

\input{supp_future.tex}

\input{supp_fullpage.tex}

\end{document}

%% file: macros.tex
\newcommand{\red}[1]{\textcolor{Red}{#1}}
\newcommand{\blue}[1]{\textcolor{Blue}{#1}}
\newcommand{\orange}[1]{\textcolor{Orange}{#1}}

\newcommand\blfootnote[1]{%
  \begingroup
  \renewcommand\thefootnote{}\footnote{#1}%
  \addtocounter{footnote}{-1}%
  \endgroup
}

\makeatletter
\renewcommand\paragraph{\@startsection{paragraph}{4}{\z@}%
                        {1ex \@plus1ex \@minus.2ex}%
                        {-1em}%
                        {\normalfont\normalsize\bfseries}}
\makeatother

\newcommand{\N}[2]{\ensuremath{\mathcal{N}(#1|#2)}}
\newcommand{\defeq}{\overset{\underset{\mathrm{def}}{}}{=}}
\newcommand{\argmin}{\operatornamewithlimits{argmin}}
\newcommand{\argmax}{\operatornamewithlimits{argmax}}

\DeclareMathOperator*{\diag}{diag}
\DeclareMathOperator*{\KL}{KL}

\def\onedot{.\xspace}
\def\etal{\emph{et~al}\onedot}
\def\ie{\emph{i.e}\onedot}

\def\bE{\mathbb{E}}

\def\cC{\mathcal{C}}

\def\cN{\mathcal{N}}
\def\cP{\mathcal{P}}

\def\ft{\mathbf{t}}
\def\bpi{\bm{\pi}}
\def\bmu{\bm{\mu}}
\def\bSig{\bm{\Sigma}}
\def\bTheta{\bm{\Theta}}

%% file: sec_abstract.tex
\begin{abstract}
Point cloud registration is a fundamental problem in 3D computer vision, graphics and robotics. For the last few decades, existing registration algorithms have struggled in situations with large transformations, noise, and time constraints. In this paper, we introduce Deep Gaussian Mixture Registration (DeepGMR), the first learning-based registration method that explicitly leverages a probabilistic registration paradigm by formulating registration as the minimization of KL-divergence between two probability distributions modeled as mixtures of Gaussians. We design a neural network that extracts pose-invariant correspondences between raw point clouds and Gaussian Mixture Model (GMM) parameters and two differentiable compute blocks that recover the optimal transformation from matched GMM parameters. This construction allows the network learn an SE(3)-invariant feature space, producing a global registration method that is real-time, generalizable, and robust to noise. Across synthetic and real-world data, our proposed method shows favorable performance when compared with state-of-the-art geometry-based and learning-based registration methods.
\keywords{Point cloud registration, Gaussian Mixture Model}
\end{abstract}

%% file: sec_introduction.tex
\section{Introduction}
\label{sec:intro}
The development of 3D range sensors~\cite{Beraldin00} has generated a massive amount of 3D data, which often takes the form of point clouds. The problem of assimilating raw point cloud data into a coherent world model is crucial in a wide range of vision, graphics and robotics applications. A core step in the creation of a world model is point cloud registration, the task of finding the transformation that aligns input point clouds into a common coordinate frame.

As a longstanding problem in computer vision and graphics, there is a large body of prior works on point cloud registration~\cite{Pomerleau2015}. However, the majority of registration methods rely solely on matching local geometry and do not leverage learned features capturing large-scale shape information. As a result, such registration methods are often \emph{local}, which means they cannot handle large transformations without good initialization. In contrast, a method is said to be \emph{global} if its output is invariant to initial poses. Several works have investigated global registration. However, they are either too slow for real-time processing \cite{yang2015go,campbell2016gogma} or require good normal estimation to reach acceptable accuracy \cite{zhou2016fast}. In many applications, such as re-localization and pose estimation, an accurate, fast, and robust global registration method is desirable.

The major difficulty of global registration lies in data association, since ICP-style correspondences based on Euclidean distances are no longer reliable. Existing registration methods provide several strategies for performing data association (see Fig.~\ref{fig:visual_comparisons}), but each of these prove problematic for global registration on noisy point clouds. Given point clouds of size $N$, DCP \cite{wang2019deep} attempts to perform point-to-point level matching like in ICP (Fig.~\ref{fig:p2p}) over all $N^2$ point pairs, which suffers from $O(N^2)$ complexity. In addition, real-world point clouds don't contain exact point-level correspondences due to sensor noise. FGR \cite{zhou2016fast} performs sparse feature-level correspondences that can be much more efficient (Fig.~\ref{fig:f2f}), but are highly dependent on the quality of features. For example, the FPFH features used in \cite{zhou2016fast} rely on consistent normal estimation, which is difficult to obtain in practice due to varying sparsity or non-rectilinear geometry~\cite{unnikrishnan_scale_2006}. Moreover, these sparse correspondences are still point-level and suffer the same problem when exact point-level correspondences don't exist. To solve this problem, one can use probabilistic methods to perform distribution-to-distribution matching (Fig.~\ref{fig:d2d}). However, the distribution parameters are not guaranteed to be consistent across different views due to the well-known problem of Gaussian Mixture \emph{non-indentifiability}~\cite{bishop2006pattern}. Thus, probabilistic algorithms like HGMR \cite{eckart2018hgmr} have only local convergence and rely on iterative techniques to update and refine point-to-distribution correspondences.

In this paper, we introduce a novel registration method that is designed to overcome these limitations by learning pose-invariant point-to-distribution parameter correspondences (Fig.~\ref{fig:ldc}).  Rather than depending on point-to-point correspondence~\cite{wang2019deep} and iterative optimization~\cite{aoki2019pointnetlk}, we solve for the optimal transformation in a single step by matching points to a probability distribution whose parameters are estimated by a neural network from the input point clouds. Our formulation is inspired by prior works on Gaussian mixture registration \cite{eckart2018hgmr,gmmreg}, but different from these works in two ways. First, our method does not involve expensive iterative procedures such as Expectation Maximization (EM)~\cite{dempster1977maximum}. Second, our network is designed to learn a consistent GMM representation across multiple point clouds rather than fit a GMM to a single reference point cloud.

Our proposed method has the following favorable properties:

\noindent \textbf{Global Registration} Point clouds can be aligned with arbitrary displacements and without any initialization. While being accurate on its own, our method can work together with local refinement methods to achieve higher accuracy.

\noindent \textbf{Efficiency} The proposed method runs on the order of 20-50 frames per second with moderate memory usage that grows linearly with the number of points, making it suitable for applications with limited computational resources.

\noindent \textbf{Robustness}
Due to its probabilistic formulation, our method is tolerant to noise and different sizes of input point clouds, and recovers the correct transformation even in the absence of exact point-pair correspondences, making it suitable for real-world scenarios.

\noindent \textbf{Differentiability}
Our method is fully differentiable and the gradients can be obtained with a single backward pass. It can be included as a component of a larger optimization procedure that requires gradients.

We demonstrate the advantages of our method over the state-of-the-art on several kinds of challenging data. The baselines consist of both recent geometry-based methods as well as learning-based methods. Our datasets contain large transformations, noise, and real-world scene-level point clouds, which we show cause problems for many state-of-the-art methods. Through its connection to Gaussian mixture registration and novel learning-based design, our proposed method performs well even in these challenging settings.

%% file: sec_related.tex
\begin{figure}[t]
    \centering
	\begin{subfigure}{0.45\linewidth}
	    \centering
        \includegraphics[width=\linewidth,trim={0pt 80pt 0pt 90pt}]{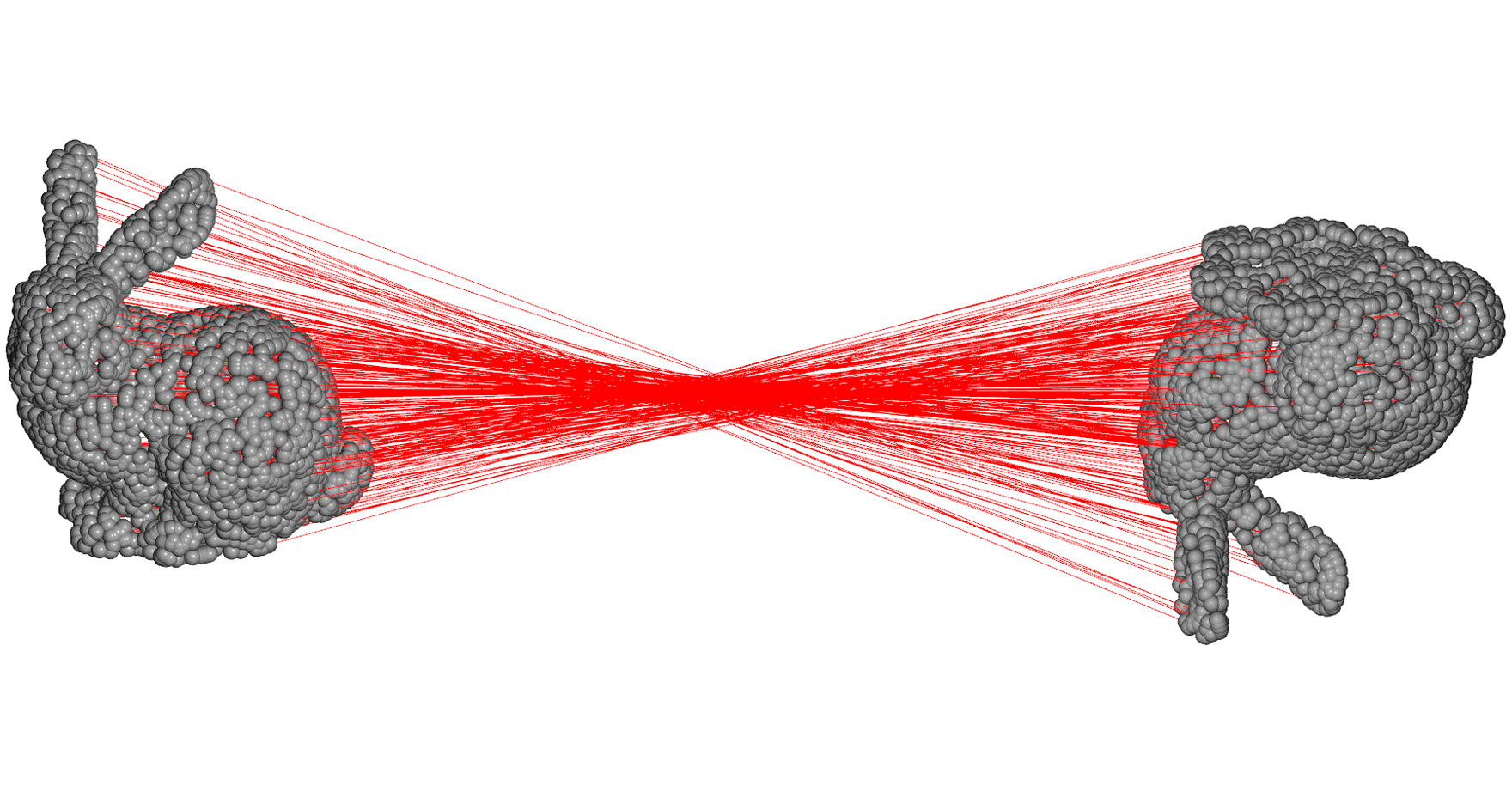}
        \caption{\scriptsize Point Matching (ICP~\cite{besl_method_1992}, DCP\cite{wang2019deep})\label{fig:p2p}}
	\end{subfigure}
	\begin{subfigure}{0.45\linewidth}
	    \centering
        \includegraphics[width=\linewidth,trim={0pt 80pt 0pt 90pt}]{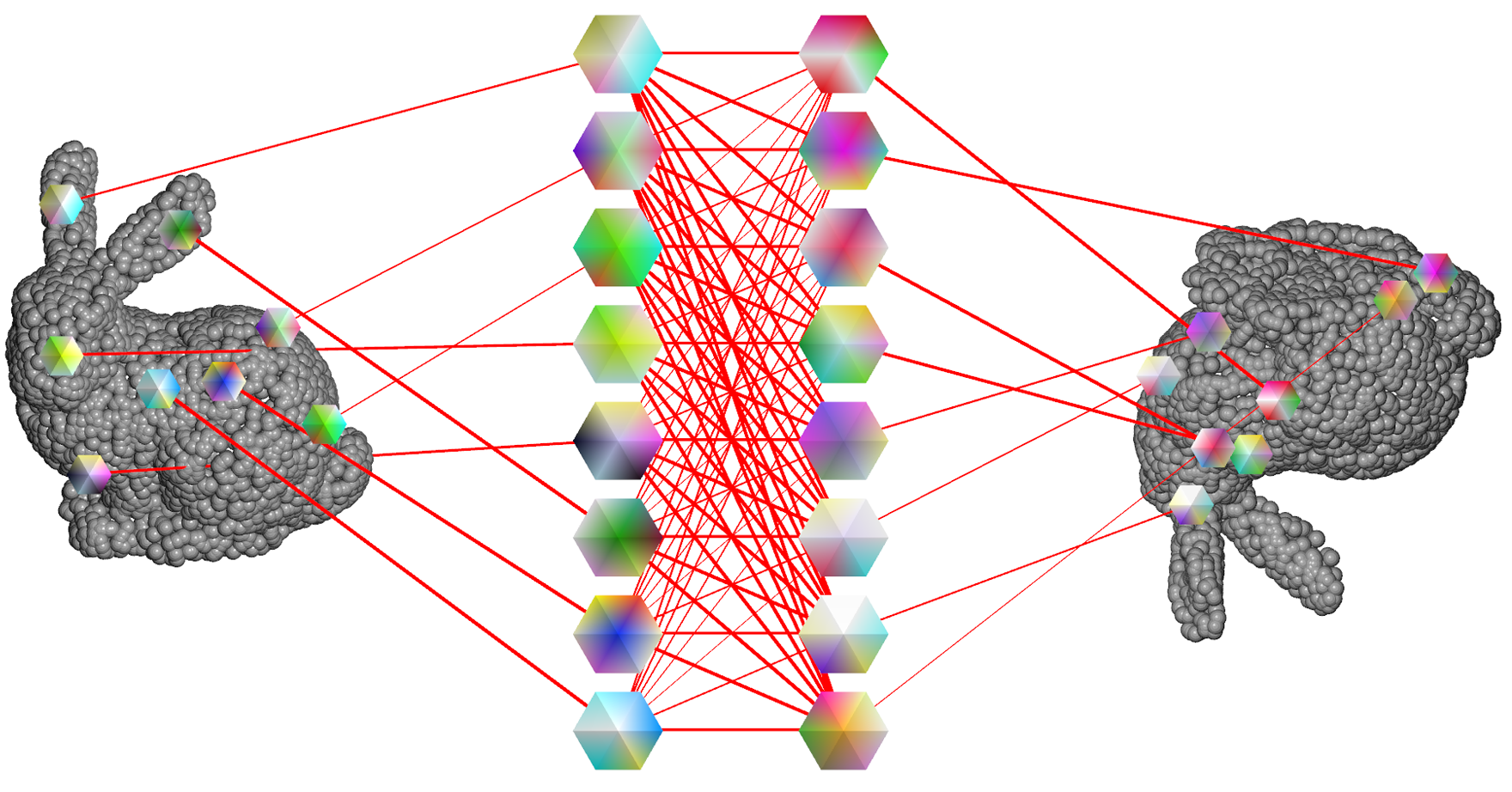}
        \caption{\scriptsize Feature Matching (FGR~\cite{zhou2016fast})\label{fig:f2f}}
	\end{subfigure} 
	\vspace{.5cm}\\
	\begin{subfigure}{0.45\linewidth}
	    \centering
        \includegraphics[width=\linewidth,trim={0pt 80pt 0pt 90pt}]{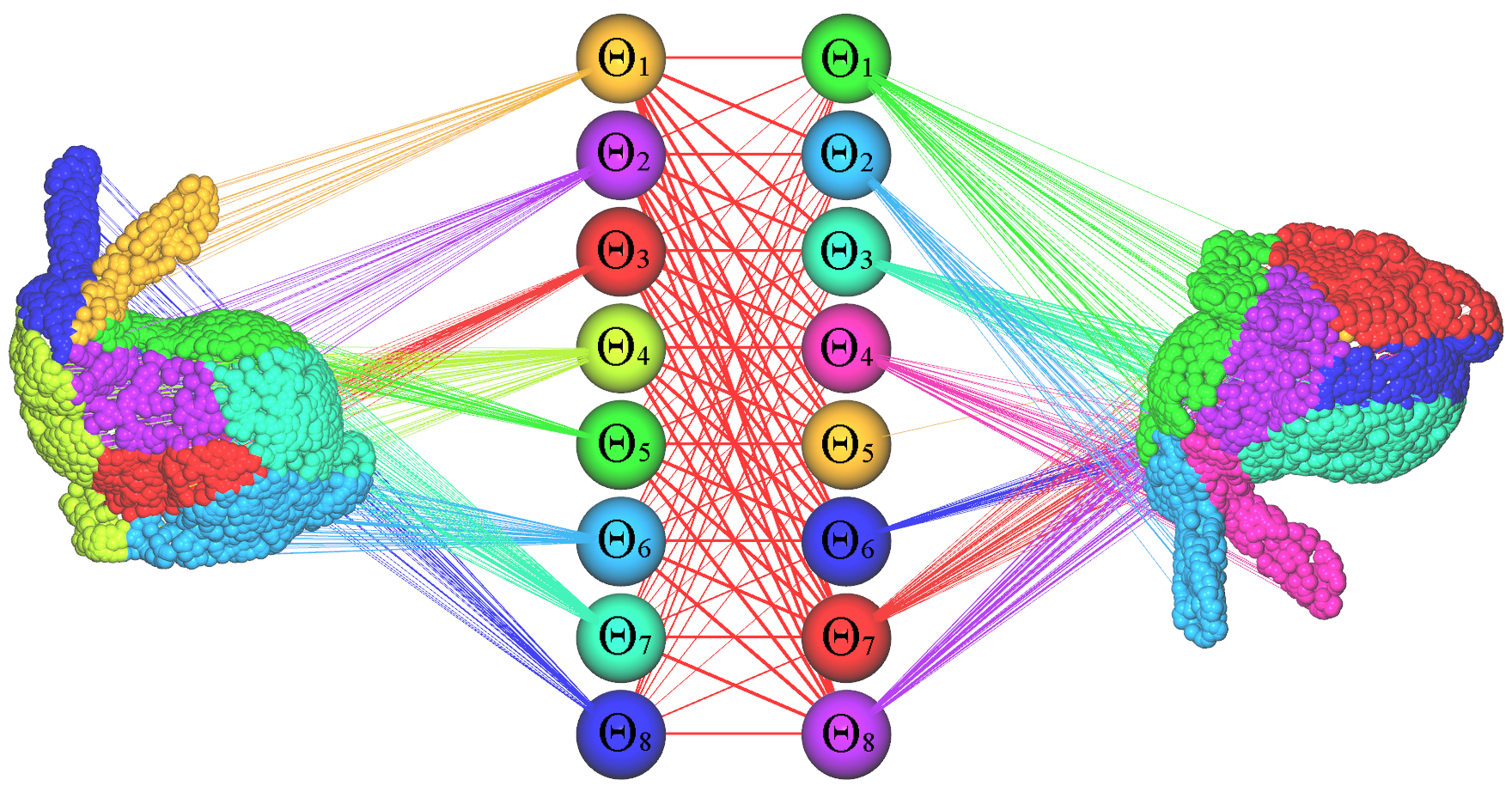}
        \caption{\scriptsize Distribution Matching (HGMR~\cite{eckart2018hgmr})\label{fig:d2d}}
	\end{subfigure}
	\begin{subfigure}{0.47\linewidth}
	    \centering
        \includegraphics[width=\linewidth,trim={0pt 80pt 0pt 90pt}]{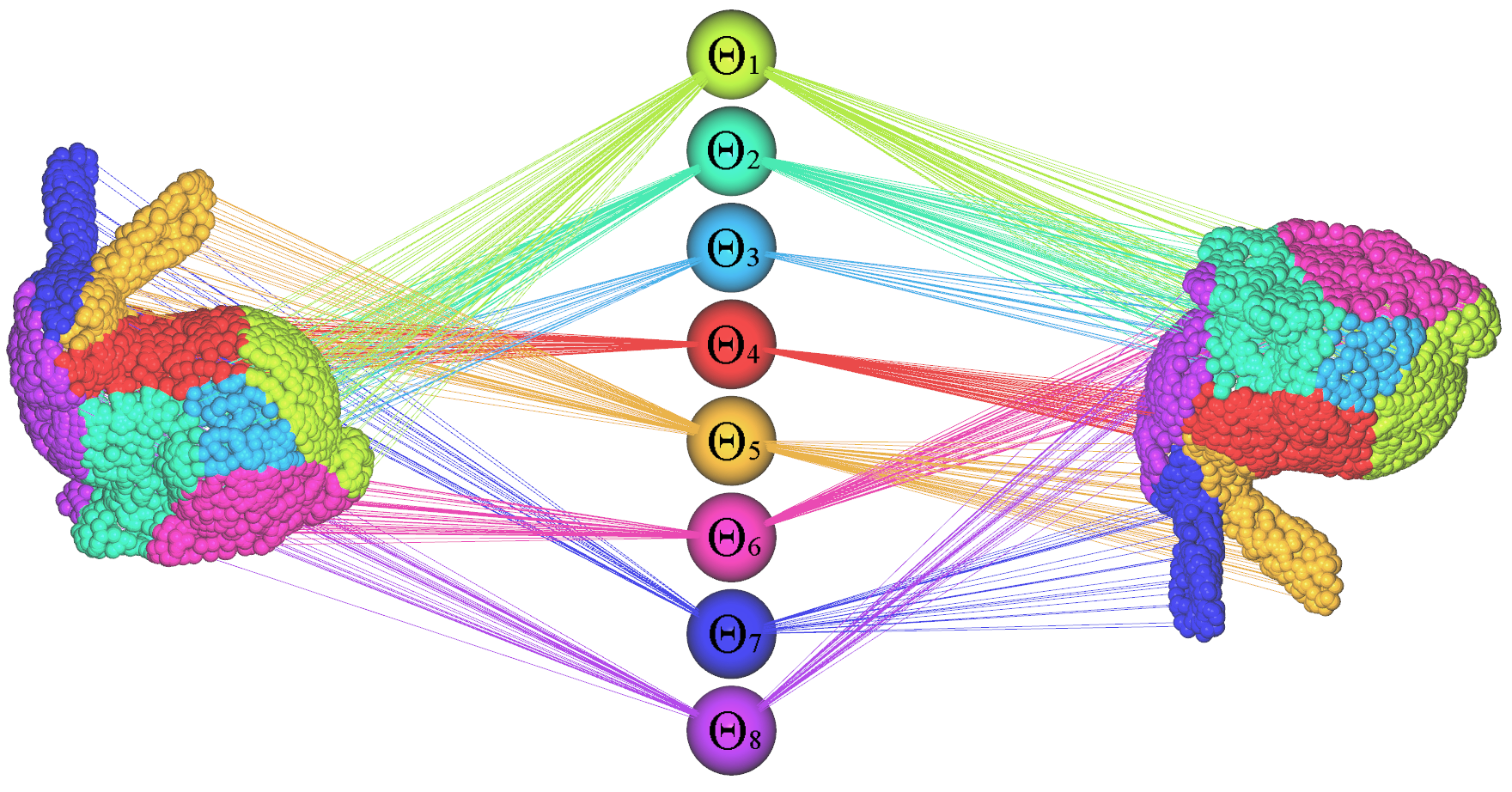}
        \caption{\scriptsize Direct Latent Correspondence  ({\bf Proposed})\label{fig:ldc}}
	\end{subfigure}
	\vspace{-.2cm}
	\caption{Comparison of data association techniques in various registration approaches. Unlike previous approaches that attempt to find point-to-point, feature-to-feature, or distribution-to-distribution level correspondences, we learn direct and pose-invariant correspondences to a distribution inside a learned latent space.}
	\vspace{-.5cm}
	\label{fig:visual_comparisons}
\end{figure}

\section{Related Work}
\label{sec:related}
Point cloud registration has remained a popular research area for many years due to its challenging nature and common presence as an important component of many 3D perception applications. Here we broadly categorize prior work into \emph{local} and \emph{global} techniques, and discuss how emerging \emph{learning-based} methods fit into these categories.

\noindent \textbf{Local Registration} Local approaches are often highly efficient and can be highly effective if limited to regimes where transformations are known \emph{a priori} to be small in magnitude. The most well-known approach is the Iterative Closest Point (ICP) algorithm~\cite{besl_method_1992,Chen92} and its many variants~\cite{rusinkiewicz2001efficient,rusinkiewicz2019symmetric}. ICP iteratively alternates between two phases: point-to-point correspondence and distance minimization. Countless strategies have been proposed for handling outliers and noise~\cite{chetverikov2005robust}, creating robust minimizers~\cite{lmicp}, or devising better distance metrics~\cite{low2004linear,gicp}.

Another branch of work on local methods concerns probabilistic registration, often via the use of GMMs and the EM algorithm~\cite{dempster1977maximum}. Traditional examples include EM-ICP~\cite{emicp}, GMMReg~\cite{gmmreg}, and methods based on the Normal Distributions Transform (NDT~\cite{ndt}). More recent examples offer features such as batch registration (JRMPC~\cite{evangelidis2017joint}) or robustness to density variance and viewing angle (DARE~\cite{jaremo2018density}). Other recent approaches have focused on efficiency, including filter-based methods~\cite{gao2019filterreg}, GPU-accelerated hierarchical methods~\cite{eckart2018hgmr}, Monte Carlo methods~\cite{dhawale2018fast} or efficient distribution-matching techniques~\cite{tabib2018manifold}. 

In our experiments, we compare our algorithm against local methods belonging to both paradigms: the Trimmed ICP algorithm with point-to-plane minimization~\cite{low2004linear,chetverikov2005robust} and Hierarchical Gaussian Mixture Registration (HGMR)~\cite{eckart2018hgmr}, a state-of-the-art probabilistic method.

\noindent \textbf{Global Registration} Unlike local methods, global methods are invariant to initial conditions, but often at the cost of efficiency. Some approaches exhaustively search $SE(3)$ via branch-and-bound techniques (Go-ICP~\cite{yang2015go}, GOGMA~\cite{campbell2016gogma} and GOSMA~\cite{campbell2019alignment}). Other approaches use local feature matching with robust optimization techniques such as RANSAC~\cite{fischler1981random} or semidefinite programming~\cite{yang2020teaser}.

One notable exception to the general rule that global methods must be inefficient is Fast Global Registration (FGR)~\cite{zhou2016fast}, which achieves invariance to initial pose while remaining as fast or faster than many local methods. We compare against FGR~\cite{zhou2016fast}, RANSAC~\cite{fischler1981random} and TEASER++~\cite{yang2020teaser} in our experiments as representatives  of state-of-the-art geometry-based global methods.

\noindent \textbf{Learning-based Registration}
Deep learning techniques on point clouds such as ~\cite{qi2017pointnet,qi2017pointnet++,su2018splatnet,wang2019dynamic} provide task-specific learned point representations that can be leveraged for robust point cloud registration.
PointNetLK~\cite{aoki2019pointnetlk}, DCP~\cite{wang2019deep} and PRNet \cite{wang2019prnet} are the closest-related registration methods to ours. PointNetLK~\cite{aoki2019pointnetlk} proposes a differentiable Lucas-Kanade algorithm~\cite{lucas1981iterative} that tries to minimize the feature distance between point clouds. DCP~\cite{wang2019deep} proposes attention-based feature matching coupled with differentiable SVD for point-to-point registration, while PRNet~\cite{wang2019prnet} uses neural networks to detect keypoints followed by SVD for final registration. However, as we show in our experiments, due to their iterative nature, PointNetLK and PRNet are local methods that do not converge under large transformations. While DCP is a global method, it performs point-to-point correspondence which suffers on noisy point clouds.

Our proposed approach can be characterized as a \emph{global} method, as well as the first \emph{learning-based}, \emph{probabilistic} registration method. To further emphasize the difference of our approach in the context of data association and matching, refer to the visual illustrations of various correspondence strategies in Fig.~\ref{fig:visual_comparisons}.

%% file: sec_preliminaries.tex
\section{GMM-Based Point Cloud Registration} 
\label{sec:preliminaries}

Before describing our approach, we will briefly review the basics of the Gaussian Mixture Model (GMM) and how it offers a maximum likelihood (MLE) framework for finding optimal alignment between point clouds, which can be solved using the Expectation Maximization (EM) algorithm~\cite{eckart2018hgmr}. We then discuss the strengths and limitations of this framework and motivate the need for learned GMMs.

A GMM establishes a multimodal generative probability distribution over 3D space ($\mathbf{x} \in \mathbb{R}^3$) as a weighted sum of $J$ Gaussian distributions,
\begin{equation}
    p(\mathbf{x} \mid \bTheta) \defeq \sum_{j=1}^J \pi_j \cN(\mathbf{x} \mid \bmu_j, \bSig_j),
\end{equation}
where $\sum_j \pi_j = 1$. GMM parameters $\bTheta$ comprise $J$ triplets $(\pi_j,\bmu_j,\bSig_j)$, where $\pi_j$ is a scalar mixture weight, $\bmu_j$ is a $3\times1$ mean vector and $\bSig_j$ is a $3\times3$ covariance matrix of the $j$-th component.

Given point clouds $\hat\cP,\cP$ and the space of permitted GMM parameterizations $\mathcal{X}$, we can formulate the registration from $\hat\cP$ to $\cP$ as a two-step optimization problem,
\begin{align}
\label{eq:mle_step1}
\mbox{\bf{Fitting}: }&\bTheta^* = \argmax_{\bTheta \in \mathcal{X}} p(\cP \mid \bTheta)\\
\label{eq:mle_step2}
\mbox{\bf{Registration}: }&T^* = \argmax_{T \in SE(3)} p(T(\hat\cP) \mid \bTheta^*)
\end{align}
where $SE(3)$ is the space of 3D rigid transformations. The fitting step fits a GMM $\bTheta^*$ to the target point cloud $\cP$, while the registration step finds the optimal transformation $T^*$ that aligns the source point cloud $\hat\cP$ to $\bTheta^*$.

Note that both steps maximize the likelihood of a point cloud under a GMM, but with respect to different parameters. In general, directly maximizing the likelihood $p$ is intractable. However, one can use EM to maximize a lower bound on $p$ by introducing a set of latent correspondence variables $\cC$. EM iterates between E-step and M-step until convergence. At each iteration $k$, the E-step updates the lower bound $q$ to the posterior over $\cC$ given a guess of the parameters $\bTheta^{(k)}$ and the M-step updates the parameters by maximizing the expected joint likelihood under $q$. As an example, the EM updates for the fitting step (Eq.~\ref{eq:mle_step1}) are as follows.
\begin{align}
\label{eq:e_step}
\mbox{\bf{E}$_{\bTheta}$: } & q(\cC) := p(\cC \mid \cP, \bTheta^{(k)})\\
\label{eq:m_step}
\mbox{\bf{M}$_{\bTheta}$: } & \bTheta^{(k+1)} := \argmax_{\bTheta \in \mathcal{X}} \bE_{q} [\ln p(\cP,\cC \mid \bTheta)]
\end{align} 
The key for EM is the introduction of latent correspondences $\cC$. Given a point cloud $\cP=\{p_i\}^{N}_{i=1}$ and a GMM $\bTheta^{(k)}=\{\pi_j,\bmu_j,\bSig_j\}_{j=1}^J$, $\cC$ comprises $NJ$ binary latent variables $\{c_{ij}\}^{N,J}_{i,j=1,1}$ whose posterior factors can be calculated as
\begin{align}
    \label{eq:cpost}
    p(c_{ij}=1\mid p_i, \bTheta^{(k)})
    &= \frac{\pi_j \N{p_i}{\bmu_j, \bSig_j}}{\sum_{j'=1}^J \pi_{j'} \N{p_i}{\bmu_{j'}, \bSig_{j'}}},
\end{align}
which can be seen as a kind of softmax over the squared Mahalanobis distances from $p_i$ to each component center $\bmu_j$. Intuitively, if $p_i$ is closer to $\bmu_j$ relative to the other components, then $c_{ij}$ is more likely to be 1. 

The introduction of $\cC$ makes the joint likelihood in the M-step (Eq.~\ref{eq:m_step}) factorizable, leading to a closed-form solution for $\bTheta^{(k+1)} \defeq \{\pi_j^*, \bmu_j^*, \bSig_j^*\}_{j=1}^J$. Let $\gamma_{ij} \defeq \bE_{q} [ c_{ij} ] $. The solution for Eq.~\ref{eq:m_step} can be written as
\begin{align}
\stepcounter{equation}
&\pi_j^* = \frac{1}{N}\sum_{i=1}^N\gamma_{ij} \label{eq:gmm_pimu}\,,
\quad\quad
N\pi_j^*\bmu_j^* =  \sum_{i=1}^N\gamma_{ij}p_i\tag{\theequation,\,\number\numexpr\theequation+1\relax}\,,\\
&N\pi_j^*\bSig_j^* =  \sum_{i=1}^N\gamma_{ij}(p_i-\bmu_j^{*})(p_i-\bmu_j^{*})^\top \label{eq:gmm_sigma}
\stepcounter{equation}
\end{align}

Likewise, the registration step (Eq.~\ref{eq:mle_step2}) can be solved with another EM procedure optimizing over the transformation parameters $T$.
\begin{align}
\label{eq:e_step_reg}
\mbox{\bf{E}$_T$: } &q(\cC) := p(\cC \mid T^{(k)}(\hat\cP), \bTheta^*)\\
\label{eq:m_step_reg}
\mbox{\bf{M}$_T$: } &T^{(k+1)} := \argmax_{T \in SE(3)} \bE_{q} [\ln p(T(\hat\cP),\cC \mid \bTheta^*)]
\end{align}

Many registration algorithms~\cite{jaremo2018density,gao2019filterreg,eckart2018hgmr,cpd,ecmpr,emicp,mlmd} can be viewed as variations of the general formulation described above. Compared to methods using point-to-point matches~\cite{Chen92,gicp,lmicp,wang2019deep}, this formulation provides a systematic way of dealing with noise and outliers using probabilistic models and is fully differentiable. However, the iterative EM procedure makes it much more computationally expensive.
Moreover, when the transformation is large, the EM procedure in Eq.~\ref{eq:e_step_reg},\ref{eq:m_step_reg} often gets stuck in local minima. This is because Eq.~\ref{eq:cpost} used in the E Step performs point-to-cluster correspondence based on locality, i.e. a point likely belongs to a component if it is close to the component's center, which leads to spurious data association between $\hat\cP$ and $\bTheta^*$ when $T$ is large. In the following section, we show how our proposed method solves the data association problem by learning pose-invariant point-to-GMM correspondences via a neural network. By doing so, we also remove the need for an iterative matching procedure.

%% file: sec_approach.tex
\begin{figure*}[t]
	\centering
    \includegraphics[width=.95\textwidth]{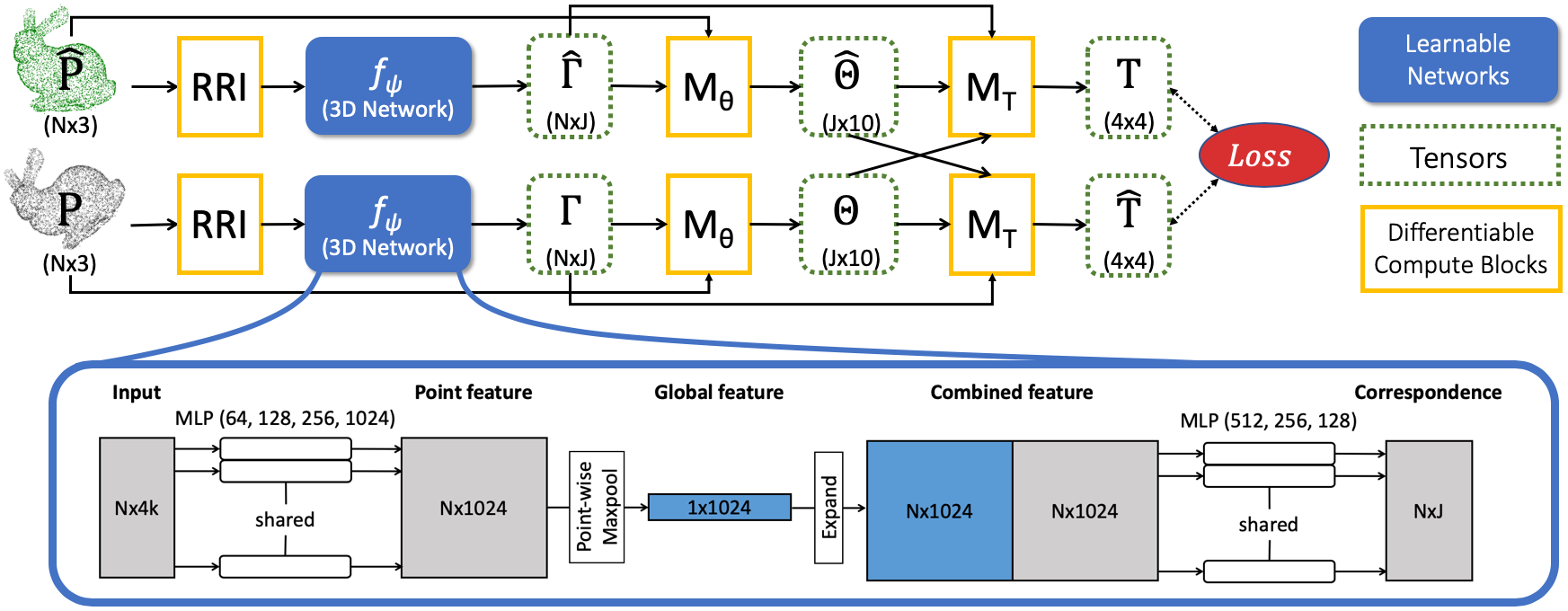}
	\caption{Given two point clouds $\hat P$ and $P$, DeepGMR estimates the optimal transformation $T$ that registers $\hat P$ to $P$ and optionally the inverse transformation $\hat T$. DeepGMR has three major components: a learnable permutation-invariant point network $f_\psi$ (Sec.~\ref{sec:gmmnet}) that estimates point-to-component correspondences $\Gamma$ and two differentiable compute blocks $\mathbf{M_{\bTheta}}$ (Sec.~\ref{sec:mtheta}) and $\mathbf{M_T}$ (Sec.~\ref{sec:mt}) that compute the GMM parameters $\bTheta$ and transformation $T$ in closed-form. We backpropagate a mean-squared loss through the differentiable compute blocks to learn the parameters of $f_\psi$.}
	\label{fig:model}
	\vspace{-.2cm}
\end{figure*}

\section{DeepGMR}
\label{sec:approach}

In this section, we give an overview of Deep Gaussian Mixture Registration (DeepGMR) (Fig.~\ref{fig:model}), the first registration method that integrates GMM registration with neural networks. DeepGMR features a correspondence network (Sec.~\ref{sec:gmmnet}) and two differentiable computing blocks (Sec.~\ref{sec:mtheta}, \ref{sec:mt}) analogous to the two EM procedures described in Sec.~\ref{sec:preliminaries}. The key idea is to replace the E-step with a correspondence network, which leads to consistent data association under large transformations, overcoming the weakness of conventional GMM registration.

\subsection{Correspondence Network $f_{\psi}$}
\label{sec:gmmnet}
The correspondence network $f_{\psi}$ takes in a point cloud with $N$ points, $\cP=\{p_i\}_{i=1}^{N}$, or a set of pre-computed features per point and produces a $N\times J$ matrix $\Gamma=\begin{bmatrix}\gamma_{ij}\end{bmatrix}$ where $\sum_{j=1}^J\gamma_{ij}=1$ for all $i$. Each $\gamma_{ij}$ represents the latent correspondence probability between point $p_i$ and component $j$ of the latent GMM.

Essentially, the correspondence network replaces the E-step of a traditional EM approach reviewed in Sec.~\ref{sec:preliminaries}. In other words, $f_{\psi}$ induces a distribution $q_{\psi}$ over latent correspondences (Eq.~\ref{eq:e_step}). Importantly, however, the learned latent correspondence matrix $\Gamma$ does not rely on Mahalanobis distances as in Eq.~\ref{eq:cpost}. This relaxation has several advantages. First, the GMMs generated by our network can deviate from the maximum likelihood estimate if it improves the performance of the downstream task. Second, by offloading the computation of $\Gamma$ to a deep neural network, higher-level contextual information can be learned in a data-driven fashion in order to produce robust non-local association predictions. Third, since the heavy lifting of finding the appropriate association is left to the neural network, only a single iteration is needed.

Finally, under this framework, the computation of $\Gamma$ can be viewed as a $J$-class point classification problem. As a result, we can leverage existing point cloud segmentation networks as our backbone. 

\subsection{$\mathbf{M_{\Theta}}$ Compute Block}
\label{sec:mtheta}

Given the outputs $\Gamma$ of $f_{\psi}$ together with the point coordinates $\cP$, we can compute the GMM parameters $\bTheta$ in closed form according to Eq.~\ref{eq:gmm_pimu}, and \ref{eq:gmm_sigma}. Since these equations are weighted combinations of the point coordinates, the estimated GMM must overlap with the input point cloud spatially, providing an effective inductive bias for learning a stable representation. We refer to the parameter-free compute block that implements the conversion from $(\Gamma, \cP) \rightarrow \bTheta$ as $\mathbf{M_{\Theta}}$.

In order to have a closed-form solution for the optimal transformation (Sec.~\ref{sec:mt}), we choose the covariances to be isotropic, \ie, $\bSig_j=\diag([\sigma_j^2,\sigma_j^2,\sigma_j^2])$. This requires a slight modification of Eq.~\ref{eq:gmm_sigma}, replacing the outer product to inner product. The number of Gaussian components $J$ remains fixed during training and testing. We choose $J=16$ in our experiments based on ablation study results (refer to the supplement for details). Note that $J$ is much smaller than the number of points, which is important for time and memory efficiency (Sec.~\ref{sec:efficiency}).

\subsection{$\mathbf{M_T}$ Compute Block}
\label{sec:mt}

In this section, we show how to obtain the optimal transformation in closed form given the GMM parameters estimated by $\mathbf{M_{\Theta}}$. We refer to this parameter-free compute block that implements the computation from $(\Gamma, \bTheta, \hat\bTheta) \rightarrow T$ as $\mathbf{M_T}$. 

We denote the source point cloud as $\hat\cP=\{\hat p_i\}$ and the target point cloud as $\cP=\{p_i\}$. The estimated source and target GMMs are denoted as $\hat\bTheta=(\hat\bpi,\hat\bmu,\hat\bSig)$ and $\bTheta=(\bpi,\bmu,\bSig)$. The latent association matrix for source and target are $\hat\Gamma=\{\hat\gamma_{ij}\}$ and $\Gamma=\{\gamma_{ij}\}$. We use $T(\cdot)$ to denote the transformation from the source to the target, consisting of rotation $R$ and translation $\ft$ (\emph{i.e.}, $T \in SE(3)$).

The maximum likelihood objective of $\mathbf{M_T}$ is to minimize the KL-divergence between the transformed latent source distribution $T(\hat\bTheta)$ and the latent target distribution~$\bTheta$, 
\begin{equation}
    T^* = \argmin_T \KL (T(\hat\bTheta) \mid \bTheta) \label{eq:kl}
\end{equation}
It can be shown that under general assumptions, the minimization in Eq.~(\ref{eq:kl}) is equivalent to maximizing the log likelihood of the transformed source point cloud $T(\hat\cP)$ under the target distribution $\bTheta$:
\begin{align}
    T^* 
    &= \argmax_T \sum_{i=1}^N \ln \sum_{j=1}^J \pi_j \cN(T(\hat p_i) \mid \bmu_j,\bSig_j) \label{eq:argmax_reg}
\end{align}
Thus, Equations~\ref{eq:kl}-\ref{eq:argmax_reg} conform to the generalized registration objective outlined in Sec.~\ref{sec:preliminaries}, Eq.~\ref{eq:m_step_reg}. We leave the proof to the supplement.

To eliminate the sum within the log function, we employ the same trick as in the EM algorithm. By introducing a set of latent variables $\cC=\{c_{ij}\}$ denoting the association of the transformed source point $T(\hat p_i)$ with component $j$ of the target GMM $\bTheta$, we can form a lower bound over the log likelihood as the expectation of the joint log likelihood $\ln p(T(\hat\cP),\cC)$ with respect to some distribution $q_{\psi}$ over $\cC$, where $\psi$ are the parameters of the correspondence network
\begin{align}
\hspace{-5pt}    T^* &= \argmax_T \bE_{q_{\psi}} [\ln p(T(\hat\cP),\cC \mid \bTheta)] \\
    &= \argmax_T \sum_{i=1}^N \sum_{j=1}^J \bE_{q_{\psi}}[c_{ij}] \ln \cN(T(\hat p_i) \mid \bmu_j,\bSig_j)
    \label{eq:mstep1}
\end{align}
Note that we parametrize $q$ with the correspondence network $f_\psi$ instead of setting $q$ as the posterior based on Mahalanobis distances (Eq.~\ref{eq:cpost}). In other words, we use $f_\psi$ to replace the $\mathbf{E}_T$ step in Eq.~\ref{eq:e_step_reg}.
Specifically, we set $\bE_{q_{\psi}}[c_{ij}]=\hat\gamma_{ij}$ and rewrite Eq. (\ref{eq:mstep1}) equivalently as the solution to the following minimization,
\begin{align}
    T^* = \argmin_T \sum_{i=1}^N \sum_{j=1}^J \hat\gamma_{ij}\|T(\hat p_i) - \bmu_j\|_{\bSig_j}^2 \label{eq:dist}
\end{align}
where $\|\cdot\|_{\bSig_j}$ denote the Mahalanobis distance.

The objective in Eq. (\ref{eq:dist}) contains $NJ$ pairs of distances which can be expensive to optimize when $N$ is large. However, we observe that the output of $\mathbf{M_\Theta}$ (\emph{i.e.}, using Eqs.~\ref{eq:gmm_pimu}, and \ref{eq:gmm_sigma}), with respect to $\hat\cP$ and $\hat\Gamma$, allows us to simplify this minimization to a single sum consisting only of the latent parameters $\hat\bTheta$ and $\bTheta$. This reduction to a single sum can be seen as a form of barycentric coordinate transformation where only the barycenters' contributions vary with respect to the transformation. Similar types of reasoning and derivations can be found in previous works using GMMs~\cite{emicp,mlmd} (see the supplement for further details). The final objective ends up with only $J$ pairs of distances, which saves a significant amount of computation. Furthermore, by using isotropic covariances in $\mathbf{M_\Theta}$, the objective simplifies from Mahalanobis distances to weighted Euclidean distances,
\begin{align}
    T^* 
    &= \argmin_T \sum_{j=1}^J \frac{\hat\pi_j}{\sigma_j^2} \|T(\hat\bmu_j) - \bmu_j\|^2\label{eq:icp_criterion}
\end{align}

Note that Eq.~\ref{eq:icp_criterion} implies a correspondence between components in the source GMM $\bTheta$ and the target GMM $\hat\bTheta$. Such a correspondence does \emph{not} naturally emerge from two independently estimated GMMs (Fig.~\ref{fig:d2d}). However, by representing the point-to-component correspondence between $\hat\cP$ and $\hat\bTheta$ and between $T(\hat\cP)$ and $\bTheta$ with the same $\Gamma$ matrix, an implicit component-to-component correspondence between $\bTheta$ and $\hat\bTheta$ is enforced, which reduces the alignment of point clouds to the alignment of component centroids (weighted by the covariances).

From Eq.~\ref{eq:icp_criterion}, we can solve $T^*$ in closed-form using a weighted version of the SVD solution in \cite{umeyama1991least} (refer to the supplement for more detail).
Note that our formulation is fundamentally different from previous learning-based approaches such as DCP~\cite{wang2019deep} that use the SVD solver on point-to-point matches. We use a rigorous probabilistic framework to reduce the registration problem to matching latent Gaussian components, which has dimension $J$ that is usually orders of magnitude smaller than the number of points $N$ (in our experiments, $J=16$ and $N=1024$). Our formulation is not only much more efficient (SVD has complexity $O(d^3)$ on a $d\times d$ matrix) but also much more robust, since exact point-to-point correspondences rarely exist in real world point clouds with sensor noise.

\subsection{Implementation} \label{sec:impl}
Our general framework is agnostic to the choice of the particular architecture of $f_\psi$. In our experiments, we demonstrate state-of-the-art results with a simple PointNet segmentation backbone~\cite{qi2017pointnet} (see inset of Fig.~\ref{fig:model}), but in principle other more powerful backbones can be easily incorporated~\cite{choy20194d,thomas2019kpconv}. Because it can be challenging for PointNet to learn rotation invariance, we pre-process the input points into a set of rigorously rotation-invariant (RRI) features, proposed by~\cite{chen2019rri}, and use these hand-crafted RRI features as the input to the correspondence network. Note that RRI features are only used in the estimation of the association matrix $\Gamma$ and not in the computation of the GMM parameters inside $\mathbf{M_{\bTheta}}$, which depends on the raw point coordinates. Our ablation studies (see supplement) show that RRI does help, but is not essential to our method.

During training, given a pair of point clouds $\hat\cP$ and $\cP$, we apply the correspondence network $f_\psi$ along with $\mathbf{M_{\bTheta}}$ and $\mathbf{M_T}$ compute blocks to obtain two transformations: $T$ from $\hat\cP$ to $\cP$ and $\hat T$ from $\cP$ to $\hat\cP$. Given the ground truth transformation $T_{gt}$ from $\hat\cP$ to $\cP$, our method minimizes the following mean-squared error:
\begin{equation}
    L = \|T T_{gt}^{-1} - I\|^2 + \|\hat T T_{gt} - I\|^2,
\end{equation}
where $T$ is the $4\times4$ transformation matrix containing both rotation and translation and $I$ is the identity matrix. We also experimented with various other loss functions, including the RMSE metric \cite{choi2015robust} used in our evaluation, but found that the simple MSE loss outperforms others by a small margin (see the supplement).

DeepGMR is fully-differentiable and non-iterative, which makes the gradient calculation more straightforward than previous iterative approaches such as PointNetLK~\cite{aoki2019pointnetlk}. We implemented the correspondence network and differentiable compute blocks using autograd in PyTorch \cite{paszke2017automatic}. For all our experiments, we trained the network for 100 epochs with batch size 32 using the Adam optimizer \cite{kingma2014adam}. The initial learning rate is 0.001 and is halved if the validation loss has not improved for over 10 epochs.

%% file: sec_results.tex
\section{Experimental Results}
\label{sec:results}

\setlist{nolistsep}
\abovedisplayskip 2pt
\belowdisplayskip 2pt

\begin{table}[t]
    \centering
    \scriptsize
    \caption{Average RMSE and recall with threshold 0.2 on various datasets. DeepGMR achieves the best performance across all datasets thanks to its ability to perform robust data association in challenging cases (Sec.~\ref{sec:analysis}). \red{Local} methods are labeled in \red{red}; \orange{Inefficient global} methods are labeled in \orange{orange} and \blue{efficient global} methods (average runtime $<1$s) are labeled in \blue{blue}. Best viewed in color.}
    \begin{tabular}{ccccccccc}
        \toprule
        \multirow{2}{*}{} & \multicolumn{2}{c}{ModelNet clean} & \multicolumn{2}{c}{ModelNet noisy} & \multicolumn{2}{c}{ModelNet unseen} & \multicolumn{2}{c}{ICL-NUIM} \\ \cmidrule(lr){2-3} \cmidrule(lr){4-5} \cmidrule(lr){6-7} \cmidrule(lr){8-9}
        & RMSE $\downarrow$ & Re@0.2 $\uparrow$ & RMSE $\downarrow$ & Re@0.2 $\uparrow$ & RMSE $\downarrow$ & Re@0.2 $\uparrow$ & RMSE $\downarrow$ & Re@0.2 $\uparrow$ \\ \midrule
        \red{ICP} \cite{Chen92} & 0.53 & 0.41 & 0.53 & 0.41 & 0.59 & 0.32 & 1.16 & 0.27 \\
        \red{HGMR} \cite{eckart2018hgmr} & 0.52 & 0.44 & 0.52 & 0.45 & 0.54 & 0.43 & 0.72 & 0.50 \\
        \red{PointNetLK} \cite{aoki2019pointnetlk} & 0.51 & 0.44 & 0.56 & 0.38 & 0.68 & 0.13 & 1.29 & 0.08 \\
        \red{PRNet} \cite{wang2019prnet} & 0.30 & 0.64 & 0.34 & 0.57 & 0.58 & 0.30 & 1.32 & 0.15 \\
        \orange{RANSAC10M+ICP}  & 0.01 & 0.99 & 0.04 & 0.96 & 0.03 & 0.98 & 0.08 & 0.98 \\
        \orange{TEASER++} \cite{yang2020teaser} & \textbf{0.00} & \textbf{1.00} & \textbf{0.01} & \textbf{0.99} & \textbf{0.01} & \textbf{0.99} & 0.09 & 0.95 \\
        \blue{RANSAC10K+ICP} & 0.08 & 0.91 & 0.42 & 0.49 & 0.30 & 0.67 & 0.17 & 0.84 \\
        \blue{FGR} \cite{zhou2016fast} & 0.19 & 0.79 & 0.2 & 0.79 & 0.23 & 0.75 & 0.15 & 0.87 \\
        \blue{DCP} \cite{wang2019deep} & 0.02 & 0.99 & 0.08 & 0.94 & 0.34 & 0.54 & 0.64 & 0.16 \\
        \blue{DeepGMR} & \textbf{0.00} & \textbf{1.00} & \textbf{0.01} & \textbf{0.99} & \textbf{0.01} & \textbf{0.99} & \textbf{0.07} & \textbf{0.99} \\
        \bottomrule
    \end{tabular}
    \label{tab:rmse}
    \vspace{-.4cm}
\end{table}

We evaluate DeepGMR on two different datasets: synthetic point clouds generated from the ModelNet40 \cite{modelnet} and real-world point clouds from the Augmented ICL-NUIM dataset~\cite{choi2015robust,Handa14icra}. Quantitative results are summarized in Table~\ref{tab:rmse} and qualitative comparison can be found in Fig.~\ref{fig:qualitative}. Additional metrics and full error distribution curves can be found in the supplement.

\paragraph{Baselines}
We compare DeepGMR against a set of strong geometry-based registration baselines, including point-to-plane ICP \cite{Chen92}, HGMR \cite{eckart2018hgmr}, RANSAC \cite{fischler1981random}, FGR \cite{zhou2016fast}\footnote{We performed a parameter search over the voxel size $v$, which is crucial for FGR's performance. We used the best results with $v=0.08$.} and TEASER++ \cite{yang2020teaser}. These methods cover the spectrum of point-based (ICP) and GMM-based (HGMR) methods, efficient global methods (FGR) as well as provably ``optimal" methods (TEASER++). We also compare against state-of-the-art learning-based methods \cite{aoki2019pointnetlk,wang2019deep,wang2019prnet}. For RANSAC, we test two variants with 10K and 10M iterations and refine the results with ICP.

\paragraph{Metrics}
Following \cite{choi2015robust,zhou2016fast}, we use the RMSE the metric designed for evaluating global registration algorithms in the context of scene reconstruction. Given a set of ground truth point correspondences $\{p_i,q_i\}_{i=1}^n$, RMSE can be computed as
\begin{equation}
    E_{RMSE}=\frac{1}{n}\sqrt{\sum_{i=1}^n\|T(p_i)-q_i\|^2} \label{eq:rmse}
\end{equation}
Since exact point correspondences rarely exist in real data, we approximate the correspondence using $\{p_i,T^*(p_i)\}_{i=1}^n$, where $T^*$ is the ground truth transformation and $p_i$ is a point sampled from the source point cloud, so Eq.~\ref{eq:rmse} becomes
\begin{equation}
    E_{RMSE}\approx\frac{1}{n}\sqrt{\sum_{i=1}^n\|T(p_i)-T^*(p_i)\|^2}
\end{equation}
In our evaluation we use $n=500$. In addition to the average RMSE, we also calculate the recall across the dataset, \ie, the percentage of instances with RMSE less than a threshold $\tau$. We followed \cite{choi2015robust} and used a threshold of $\tau=0.2$.

\subsection{Datasets}
\paragraph{Synthetic data}
We generated 3 variations of synthetic point cloud data for registration from ModelNet40 \cite{modelnet}, a large repository of CAD models. Each dataset follows the train/test split in \cite{modelnet} (9843 for train/validation and 2468 for test across 40 object categories).

\emph{ModelNet Clean}. 
This dataset contains synthetic point cloud pairs without noise, so exact point-to-point correspondences exist between the point cloud pairs. Specifically, we sample 1024 points uniformly on each object model in ModelNet40 and apply two arbitrary rigid transformations to the same set of points to obtain the source and target point clouds. Note that similar setting has been used to evaluate learning-based registration methods in prior works \cite{aoki2019pointnetlk,wang2019deep}, but the rotation magnitude is restricted, whereas we allow \emph{unrestricted} rotation.

\emph{ModelNet Noisy}
The setting in \emph{ModelNet clean} assumes exact correspondences, which is unrealistic in the real world. In this dataset, we perturb the sampled point clouds with Gaussian noise sampled from $\cN(0,0.01)$. Note that unlike prior works \cite{aoki2019pointnetlk,wang2019deep}, we add noise independently to the source and target, breaking exact correspondences between the input point clouds.

\emph{ModelNet Unseen}
Following \cite{wang2019deep,wang2019prnet}, we generate a dataset where train and test point clouds come from different object categories to test each method's generalizability. Specifically, we train our method using point clouds from 20 categories and test on point clouds from the held-out 20 categories unseen during training. The point clouds are the same as in \emph{ModelNet noisy}.

\paragraph{Real-world Data}
To demonstrate applicability on real-world data, we tested each method using the point clouds derived from RGB-D scans in the Augmented ICL-NUIM dataset \cite{choi2015robust}. However, the original dataset contains only 739 scan pairs which are not enough to train and evaluate learning-based methods (a network can easily overfit to a small number of fixed transformations). To remedy this problem, we take every scan available and apply the same augmentation as we did for the ModelNet data, \ie, resample the point clouds to 1024 points and apply two arbitrary rigid transformations. In this way, our augmented dataset contains both point clouds with realistic sensor noise and arbitrary transformation between input pairs. We split the dataset into 1278 samples for train/validation and 200 samples for test.

\subsection{Analysis} \label{sec:analysis}
\begin{figure*}[t]
    \centering
    \tabcolsep 0pt
    \scriptsize
    \begin{tabular}{cccccccc}
        Input & ICP\cite{Chen92} & HGMR\cite{eckart2018hgmr} & PointNetLK \cite{aoki2019pointnetlk} & PRNet\cite{wang2019prnet} & FGR\cite{zhou2016fast} & DCP\cite{wang2019deep} & DeepGMR \\
        \includegraphics[width=0.12\linewidth]{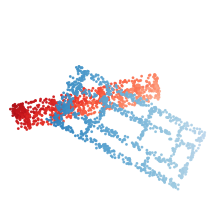} & \includegraphics[width=0.12\linewidth]{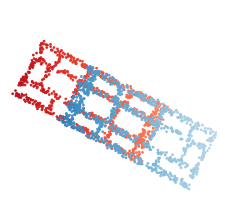} & \includegraphics[width=0.12\linewidth]{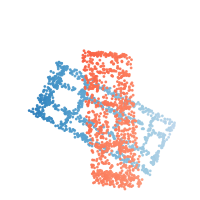} & \includegraphics[width=0.12\linewidth]{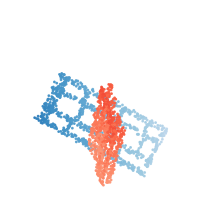} & \includegraphics[width=0.12\linewidth]{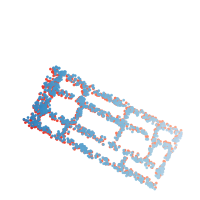} & \includegraphics[width=0.12\linewidth]{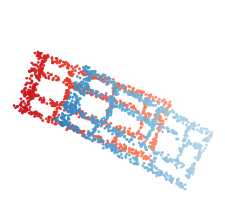} & \includegraphics[width=0.12\linewidth]{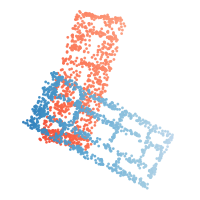} & \includegraphics[width=0.12\linewidth]{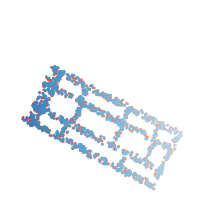} \\
        0.824 & 1.317 & 1.086 & 0.922 & 0.024 & 0.643 & 1.296 & 0.003 \\
        \includegraphics[width=0.12\linewidth]{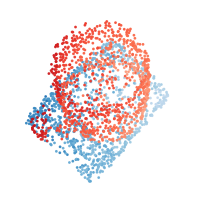} & \includegraphics[width=0.12\linewidth]{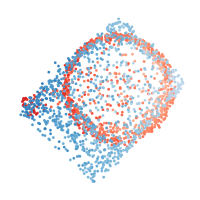} & \includegraphics[width=0.12\linewidth]{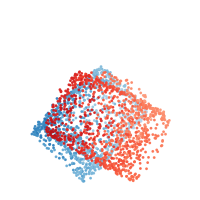} & \includegraphics[width=0.12\linewidth]{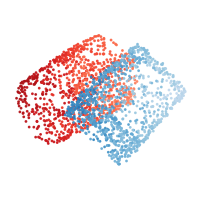} & \includegraphics[width=0.12\linewidth]{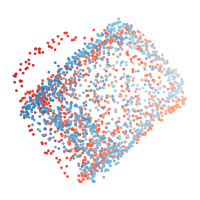} & \includegraphics[width=0.12\linewidth]{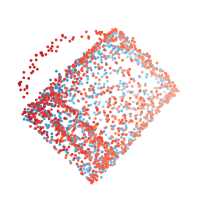} & \includegraphics[width=0.12\linewidth]{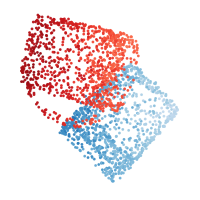} & \includegraphics[width=0.12\linewidth]{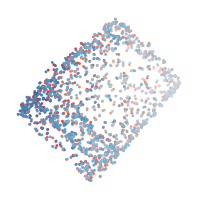} \\
        1.195 & 0.988 & 1.557 & 1.250 & 0.285 & 0.419 & 1.480 & 0.005 \\
        \includegraphics[width=0.12\linewidth]{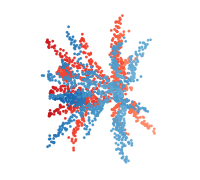} & \includegraphics[width=0.12\linewidth]{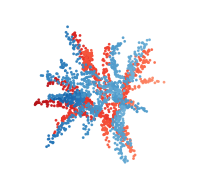} & \includegraphics[width=0.12\linewidth]{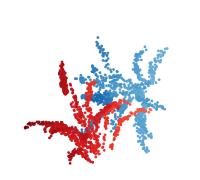} & \includegraphics[width=0.12\linewidth]{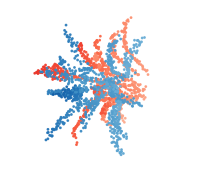} & \includegraphics[width=0.12\linewidth]{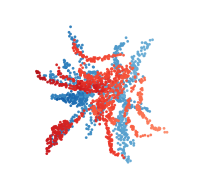} & \includegraphics[width=0.12\linewidth]{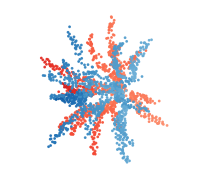} & \includegraphics[width=0.12\linewidth]{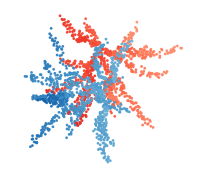} & \includegraphics[width=0.12\linewidth]{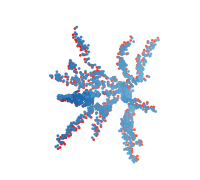} \\
        0.751 & 0.714 & 0.965 & 1.021 & 0.699 & 0.753 & 0.878 & 0.012 \\
        \includegraphics[width=0.12\linewidth]{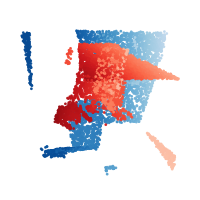} & \includegraphics[width=0.12\linewidth]{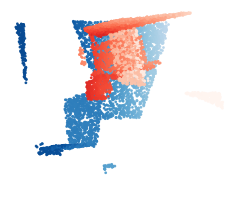} & \includegraphics[width=0.12\linewidth]{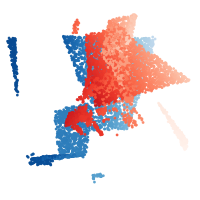} & \includegraphics[width=0.12\linewidth]{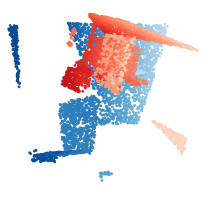} & \includegraphics[width=0.12\linewidth]{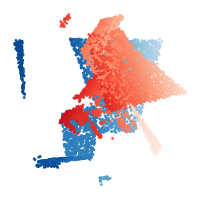} & \includegraphics[width=0.12\linewidth]{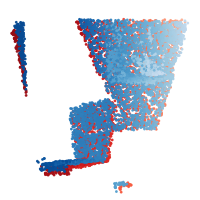} & \includegraphics[width=0.12\linewidth]{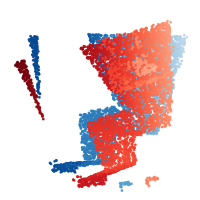} & \includegraphics[width=0.12\linewidth]{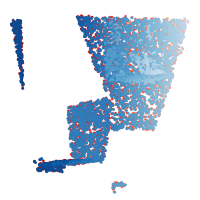} \\
        1.696 & 1.910 & 1.836 & 1.814 & 1.818 & 0.050 & 0.276 & 0.005 \\
        \includegraphics[width=0.12\linewidth]{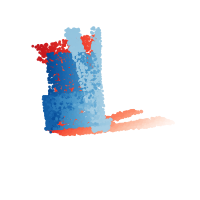} & \includegraphics[width=0.12\linewidth]{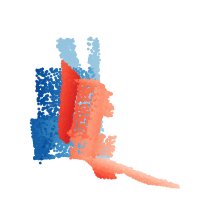} & \includegraphics[width=0.12\linewidth]{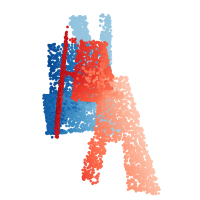} & \includegraphics[width=0.12\linewidth]{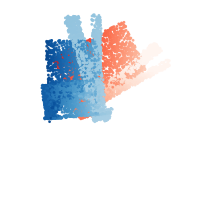} & \includegraphics[width=0.12\linewidth]{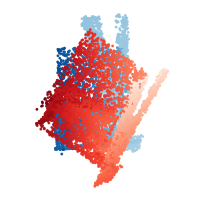} & \includegraphics[width=0.12\linewidth]{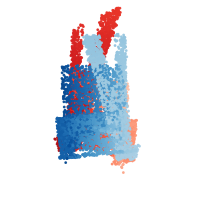} & \includegraphics[width=0.12\linewidth]{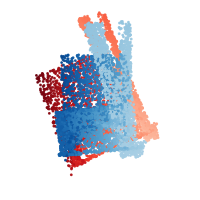} & \includegraphics[width=0.12\linewidth]{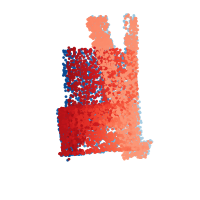} \\
        1.432 & 1.735 & 1.598 & 1.125 & 0.710 & 1.635 & 0.301 & 0.038
    \end{tabular}
    \vspace{-.2cm}
    \caption{Qualitative registration results on noisy ModelNet40 (top 3 rows) and ICL-NUIM point clouds (bottom 2 rows). The RMSE of each example is labeled below the plot. These examples highlight some typical failure modes of existing methods such as 1) ignoring parts with sparse point samples 2) erroneous data association due to repetitive structures and symmetry. DeepGMR avoids these errors by estimating consistent point-to-distribution correspondences and performing robust registration on learned GMMs.}
    \label{fig:qualitative}
    \vspace{-.6cm}
\end{figure*}

From the quantitative results in Table~\ref{tab:rmse}, we can see that DeepGMR consistently achieves good performance across challenging scenarios, including noise, unseen geometry and real world data. TEASER++, the only baseline that is able to match the performance of DeepGMR on synthetic data, is over 1000 times slower (13.3s per instance) than DeepGMR (11ms per instance).
In the rest of this section, we analyze common failure modes of the baselines and explain why DeepGMR is able to avoid these pitfalls. Typical examples can be found in the qualitative comparison shown in Fig.~\ref{fig:qualitative}.

\paragraph{Repetitive structures and symmetry}
Many point clouds contains repetitive structures (e.g. the bookshelf in the first row of Fig.~\ref{fig:qualitative}) and near symmetry (e.g. the room scan in the last row of Fig.~\ref{fig:qualitative}). This causes confusion in the data association for methods relying on distance-based point matching (e.g. ICP) or local feature descriptors (e.g. FGR). Unlike these baselines, DeepGMR performs data association globally. With the learned correspondence, a point can be associated to a distant component based on the context of the entire shape rather than a local neighborhood.

\paragraph{Parts with sparse point samples}
Many thin object parts (e.g. handle of the mug in the second row in Fig.~\ref{fig:qualitative}) are heavily undersampled but are crucial for identifying correct alignments. These parts are easily ignored as outliers in an optimization whose objective is to minimize correspondence distances. Nevertheless, by doing optimization on a probabilistic model learned from many examples, DeepGMR is able to recover the correct transformation utilizing features in sparse regions.

\paragraph{Irregular geometry}
Among all the methods, FGR is the only one that performs significantly better on ICL-NUIM data than ModelNet40. This can be attributed to the FPFH features used by FGR to obtain initial correspondences. FPFH relies on good point normal estimates, and since the indoor scans of ICL-NUIM contain mostly flat surfaces with corners (e.g. the fourth row of Fig.~\ref{fig:qualitative}), the normals can be easily estimated from local neighborhoods. In ModelNet40, however, many point clouds have irregular geometry (e.g. the plant in the third row of Fig.~\ref{fig:qualitative}) which makes normal estimates very noisy. As a result, FGR's optimization fails to converge due to the lack of initial inlier correspondences.

\begin{table}[t]
    \centering
    \scriptsize
    \tabcolsep 3pt
    \caption{Average running time (ms) of efficient registration methods on ModelNet40 test set. DeepGMR is significantly faster than other learning based method \cite{aoki2019pointnetlk,wang2019deep,wang2019prnet} and comparable to geometry-based methods designed for efficiency \cite{eckart2018hgmr,zhou2016fast}. Baselines not listed (RANSAC10M+ICP, TEASER++) have running time on the order of 10s. OOM means a 16GB GPU is out of memory with a forward pass on a single instance.}
    \vspace{.2cm}
    \begin{tabular}{ccccccccc}
        \toprule
        \# points & ICP & HGMR & PointNetLK & PRNet & RANSAC10K+ICP & FGR & DCP & DeepGMR \\ \midrule
        1000 & 184 & 33 & 84 & 153 & 95 & 22 & 67 & \textbf{11} \\
        2000 & 195 & 35 & 90 & 188 & 101 & 32 & 90 & \textbf{19} \\
        3000 & 195 & 37 & 93 & OOM & 113 & 37 & 115 & \textbf{26} \\
        4000 & 198 & 39 & 106 & OOM & 120 & 40 & 135 & \textbf{34} \\
        5000 & 201 & \textbf{42} & 109 & OOM & 124 & \textbf{42} & 157 & 47 \\
        \bottomrule
    \end{tabular}
    \label{tab:time}
    \vspace{-.5cm}
\end{table}

\vspace{-.3cm}
\subsection{Time Efficiency} \label{sec:efficiency}
We compute the average amount of time required to register a single pair of point clouds versus the point cloud size across all test instances in the ModelNet40. The results are shown in Table~\ref{tab:time}. Two baselines, RANSACK10M+ICP and TEASER++, are not listed because their average running time is over 10s for point clouds with 1000 points. ICP \cite{Chen92}, FGR \cite{zhou2016fast} and RANSAC10K+ICP are tested on an AMD RYZEN 7 3800X 3.9 GHz CPU. HGMR \cite{eckart2018hgmr}, PointLK \cite{aoki2019pointnetlk} and DeepGMR are tested on a NVIDIA RTX 2080 Ti GPU. DCP \cite{wang2019deep} and PRNet \cite{wang2019prnet} are tested on a NVIDIA Tesla V100 GPU because they require significant GPU memory. It can be seen that DeepGMR is the fastest among all methods thanks to its non-iterative nature and the reduction of registration to the task of learning correspondences to a low-dimensional latent probability distribution. 

%% file: sec_conclusion.tex
\section{Conclusion}
\label{sec:conclusion}
We have proposed DeepGMR, a first attempt towards learning-based probabilistic registration. Our experiments show that DeepGMR outperforms state-of-the-art geometry-based and learning-based registration baselines across a variety of data settings. Besides being robust to noise, DeepGMR is also generalizable to new object categories and performs well on real world data. We believe DeepGMR can be useful for applications that require accurate and efficient global alignment of 3D data, and furthermore, its design provides a novel way to integrate 3D neural networks inside a probabilistic registration paradigm.

%% file: supp_overview.tex
\section{Overview}

In this document, we provide additional details, discussions, and experiments to support the original submission. Below is a summary of the contents.
\begin{itemize}
    \item Sec.~\ref{supp:math} provides detailed derivation and proofs for the $\mathbf{M_T}$ compute block described in Sec.~4.3 of the main paper.
    \item Sec.~\ref{supp:quant} contains auxiliary results, including error distribution curves (Fig.~\ref{fig:cdf}), ablation studies (Fig.~\ref{fig:abl}), robustness tests (Fig.~\ref{fig:robust}) and category-specific results (Table~\ref{tab:cat}) in support of the major results in Sec.~5 of the main paper.
    \item Sec.~\ref{supp:qual} shows additional visualizations of registration results (Fig.~\ref{fig:qual}) and the learned latent GMMs (Fig.~\ref{fig:gmm}).
    \item Sec.~\ref{supp:future} discusses a limitation of our method and suggests directions for future research.
\end{itemize}

%% file: supp_math.tex
\section{Additional Derivation} \label{supp:math}

\subsection{KL-divergence to Maximum Likelihood}
We prove the conditions whereby Eq.~(12) in the main paper is equivalent to Eq.~(13), \ie the conditions under which minimizing the KL-divergence from the transformed source distribution $T(\hat\bTheta)$ to the target distribution $\bTheta$ is equivalent to maximizing the likelihood of the transformed source point cloud $T(\hat\cP)$ under the target distribution $\bTheta$.

Given two probability distributions $p(x)$ and $q(x)$ on $\mathcal{X}$, the KL-divergence between $p$ and $q$ is defined to be
\begin{align}
    \KL{(p\mid q)} &= \int_{\mathcal{X}} p(x) \left[\ln p(x) - \ln q(x)\right] dx \\
    &= \bE_{x \sim p(x)}\ln p(x) - \bE_{x \sim p(x)}\ln q(x)
\end{align}

Thus, we can write the KL-minimization problem in Eq.~(12) of the main paper as follows,
\begin{align}
    T^* &= \argmin_T \KL{(T(\hat\bTheta) \mid \bTheta)} \\
    &= \argmin_T \bE_{x \sim p(x \mid T(\hat\bTheta))} \ln p(x \mid T(\hat\bTheta)) - \bE_{x \sim p(x \mid T(\hat\bTheta))} \ln p(x \mid \bTheta)
\end{align}

Note that the first term, the negative entropy of $p(x|T(\hat\bTheta))$, is invariant with respect to $T$, so we end up with
\begin{equation}
    T^* = \argmax_T \bE_{x \sim p(x \mid T(\hat\bTheta))} \ln p(x \mid \bTheta) \label{eq:max_ln}
\end{equation}

Thus, minimizing the KL-divergence from $T(\hat\bTheta)$ to $\bTheta$ is equivalent to maximizing the expected log likelihood of data distributed according to $T(\hat\bTheta)$ under $\bTheta$. Or, in other words, minimizing the cross-entropy. Note that the transformed source point cloud $T(\hat\cP)=\{T(\hat p_i)\}_{i=1}^N$ are sampled \emph{iid} from the distribution $p(x|T(\hat\bTheta))$. Using the law of large numbers, given a suitably large point cloud, we can approximate the expectation in Eq.~(\ref{eq:max_ln}) as
\begin{align}
    \argmax_T \bE_{x \sim p(x|T(\hat\bTheta))} \ln p(x|\bTheta) &\approx \argmax_T \frac{1}{N} \sum_{i=1}^N \ln p(T(\hat p_i) \mid \bTheta) \\
    &= \argmax_T \sum_{i=1}^N \ln \sum_{j=1}^J \pi_j \cN(T(\hat p_i) \mid \bmu_j,\bSig_j))
\end{align}
which gives us Eq.~(13) in the main paper.

\subsection{Single Sum Reduction}
We show how to reduce the $NJ$ pairs of distances in Eq.~(16) of the main paper to $J$ pairs of distances in Eq.~(17) of the main paper using the output of $\mathbf{M_{\bTheta}}$ (Eqs.~(7,8,9) in the main paper).

The calculations inside $\mathbf{M_{\bTheta}}$ (Eqs.~(7,8,9) in the main paper) determine the relationships between the correspondence matrix $\hat\Gamma=\{\hat\gamma_{ij}\}_{i,j=1,1}^{N,J}$, the point coordinates $\hat\cP=\{\hat p_{i}\}_{i=1}^N$ and the GMM parameters $\hat\bTheta=\{\hat\pi_j,\hat\bmu_j,\hat\bSig_j\}_{j=1}^J$. Specifically, we can rewrite Eqs.~(7,8) in the main paper as
\begin{align}
    \sum_{i=1}^N \hat\gamma_{ij} &= N\hat\pi_j \label{eq:gamma_pi} \\ 
    \sum_{i=1}^N \hat\gamma_{ij} T(\hat p_i) &= N\hat\pi_j T(\hat\bmu_j) \label{eq:gamma_mu}
\end{align}

To prove the latter identity, note that the 3D rigid transformation $T$ is a linear operator. Therefore,
\begin{align}
    \sum_{i=1}^N \hat\gamma_{ij} T(\hat p_i) &= T \left( \sum_{i=1}^N \hat\gamma_{ij} \hat p_i \right)\\
    &= T( N\hat{\pi}_j \hat\bmu_j )\\
    &= N\hat{\pi}_j T(\hat\bmu_j)
\end{align}

Next, we expand the right hand side of Eq.~(16) in the main paper, which contains $NJ$ pairs of distances, using Eqs.~(\ref{eq:gamma_pi},\ref{eq:gamma_mu}).
\begin{align}
    & \sum_{i=1}^N \sum_{j=1}^J \hat\gamma_{ij} \|T(\hat p_i) - \bmu_j\|_{\bSig_j}^2 \\
    =& \sum_{j=1}^J\sum_{i=1}^N \hat\gamma_{ij}\|T(\hat p_i)\|_{\bSig_j}^2 - 2\sum_{j=1}^J \bmu_j^\top \bSig_j^{-1} \sum_{i=1}^N \hat\gamma_{ij} T(\hat p_i) + \sum_{j=1}^J \|\bmu_j^\top\|_{\bSig_j}^2 \sum_{i=1}^N \hat\gamma_{ij} \\
    =& \sum_{j=1}^J \left( \sum_{i=1}^N \hat\gamma_{ij}\|T(\hat p_i)\|_{\bSig_j}^2 - 2 \bmu_j^\top \bSig_j^{-1} N\hat{\pi}_j T(\hat\bmu_j) + \|\bmu_j^\top\|_{\bSig_j}^2 N\hat\pi_j \right) \label{eq:intermediate}
\end{align}

Now, we complete the square by adding $N\hat{\pi}_j\|T(\hat\bmu_j)\|_{\bSig_j}^2$ to the latter two terms in the summation and subtracting it from the first term. For the latter two terms, we have
\begin{align}
    & N\hat{\pi}_j\|T(\hat\bmu_j)\|_{\bSig_j}^2 - 2 \bmu_j^\top \bSig_j^{-1} N\hat{\pi}_j T(\hat\bmu_j) + \|\bmu_j^\top\|_{\bSig_j}^2 N\hat\pi_j \\
    =& N\hat{\pi}_j \left( \|T(\hat\bmu_j)\|_{\bSig_j}^2 - 2 \bmu_j^\top \bSig_j^{-1} T(\hat\bmu_j) + \|\bmu_j^\top\|_{\bSig_j}^2 \right) \\
    =& N\hat{\pi}_j \|\bmu_j - T(\hat\bmu_j)\|_{\bSig_j}^2 \label{eq:reduced}
\end{align}

For the first term, we have
\begin{align}
& \sum_{i=1}^N \hat\gamma_{ij}\|T(\hat p_i)\|_{\bSig_j}^2 - N\hat{\pi}_j\|T(\hat\bmu_j)\|_{\bSig_j}^2 \\
=& \sum_{i=1}^N \hat\gamma_{ij}\|T(\hat p_i)\|_{\bSig_j}^2 - 2N\hat{\pi}_j\|T(\hat\bmu_j)\|_{\bSig_j}^2 + N\hat{\pi}_j\|T(\hat\bmu_j)\|_{\bSig_j}^2 \\
=& \sum_{i=1}^N \hat\gamma_{ij}\|T(\hat p_i)\|_{\bSig_j}^2 - 2N\hat{\pi}_j T(\hat\bmu_j)^\top \bSig_j^{-1} T(\hat\bmu_j) + \sum_{i=1}^N \hat\gamma_{ij} \|T(\hat\bmu_j)\|_{\bSig_j}^2 \label{eq:square_step1} \\
=& \sum_{i=1}^N \hat\gamma_{ij}\|T(\hat p_i)\|_{\bSig_j}^2 - \sum_{i=1}^N \hat\gamma_{ij} T(\hat p_j)^\top \bSig_j^{-1} T(\hat\bmu_j) + \sum_{i=1}^N \hat\gamma_{ij} \|T(\hat\bmu_j)\|_{\bSig_j}^2 \label{eq:square_step2} \\
=& \sum_{i=1}^N \hat\gamma_{ij} \|T(\hat p_i) - T(\hat\bmu_j)\|_{\bSig_j}^2 \label{eq:invariant}
\end{align}
Eq.~(\ref{eq:square_step1}) and (\ref{eq:square_step2}) uses the relationship in Eq.~(\ref{eq:gamma_pi}) and (\ref{eq:gamma_mu}) respectively. Notice that the result in Eq.~(\ref{eq:invariant}) is invariant to $T$ because we assume $\Sigma_j$ is isotropic. Therefore, if we are optimizing over $T$, we can reduce Eq.~(\ref{eq:intermediate}), \ie the right hand side of Eq.~(16) in the main paper, to the single term in Eq.~(\ref{eq:reduced}), which gives us
\begin{align}
T^* &= \argmin_T \sum_{i=1}^N \sum_{j=1}^J \hat\gamma_{ij} \|T(\hat p_i) - \bmu_j\|_{\bSig_j}^2 \\
&= \argmin_T \sum_{j=1}^J N\hat\pi_j \|\bmu_j - T(\hat\bmu_j)\|_{\bSig_j}^2 \\
&= \argmin_T \sum_{j=1}^J \frac{\hat\pi_j}{\sigma_j^2} \|T(\hat\bmu_j) - \bmu_j\|^2
\end{align}
This is exactly Eq.~(17) in the main paper.

\subsection{SVD Solution}
We derive the solution to the weighted ICP criterion in Eq.~(17) of the main paper using a weighted version of Umeyama's method~\cite{umeyama1991least}. First, we center the data and construct the cross-covariance matrix $M$,
\begin{align}
\bmu_c &\defeq \sum_{j=1}^J \hat{\pi}_j\bmu_j \\
\hat\bmu_c &\defeq \sum_{j=1}^J \hat\pi_j \hat\bmu_j \\
M &\defeq \sum_{j=1}^J \frac{\hat\pi_j}{\sigma_j^2} (\bmu_j - \bmu_c) (\hat\bmu_j - \hat\bmu_c)^T
\end{align}

Assuming $T \in SE(3)$, given the SVD decomposition of $M = USV^T$, the optimal rotation $R^*$ and translation $\ft^*$ are as follows,
\begin{align}
\label{eq:MLE_R}
R^* &= V \begin{bmatrix} 1 & & 0 & 0\\ 0 & & 1 & 0 \\ 0 & & 0 & \det{VU^T} \end{bmatrix}U^T \\
\ft^* &= \bmu_c - R^*\hat\bmu_c
\end{align}

The center matrix in Equation~(\ref{eq:MLE_R}) comes from the fact that we want to enforce $\det{R^*} = +1$ to prevent reflections.

%% file: supp_quant.tex
\section{Additional Quantitative Results} \label{supp:quant}
\subsection{Full Error Distribution}
\begin{figure}[h]
    \centering
	\begin{subfigure}{0.45\linewidth}
	    \centering
        \includegraphics[width=\linewidth]{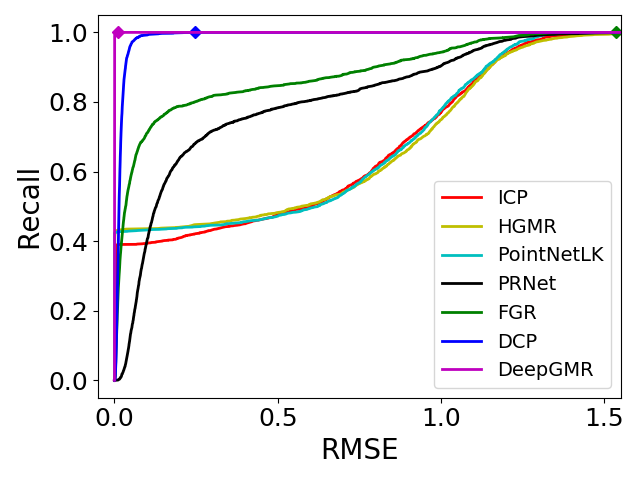}
        \vspace*{-7mm}
	    \caption{ModelNet clean}
	\end{subfigure}
	\begin{subfigure}{0.45\linewidth}
	    \centering
        \includegraphics[width=\linewidth]{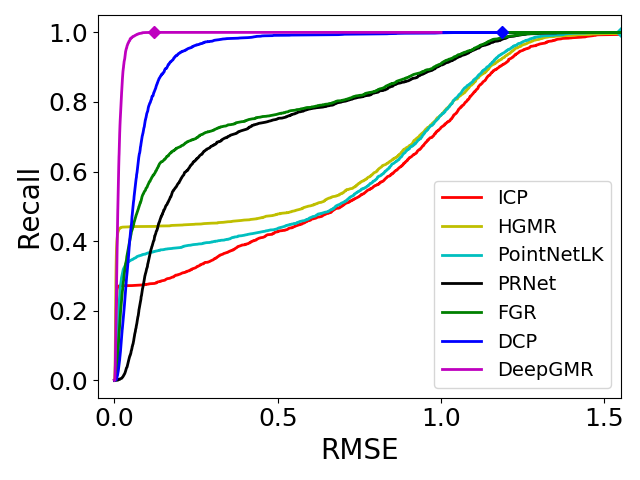}
        \vspace*{-7mm}
	    \caption{ModelNet noisy}
	\end{subfigure}
	\vspace{1mm} \\
	\begin{subfigure}{0.45\linewidth}
	    \centering
        \includegraphics[width=\linewidth]{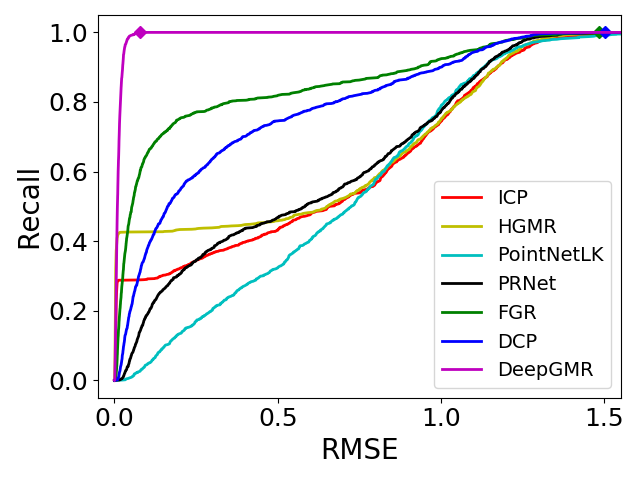}
        \vspace*{-7mm}
	    \caption{ModelNet unseen}
	\end{subfigure}
	\begin{subfigure}{0.45\linewidth}
	    \centering
        \includegraphics[width=\linewidth]{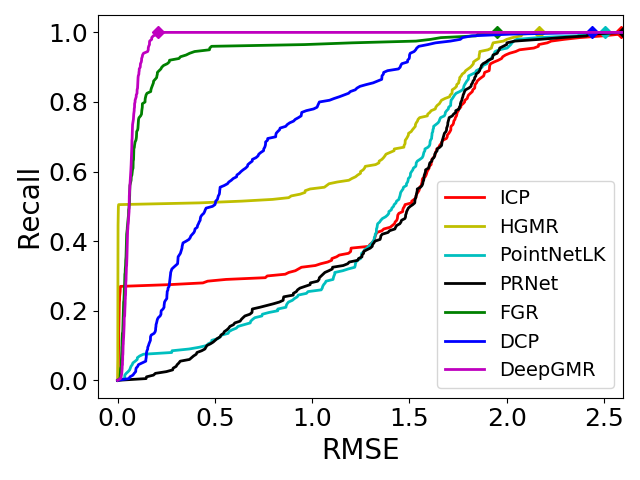}
        \vspace*{-7mm}
	    \caption{ICL-NUIM}
	\end{subfigure}
    \vspace*{-2mm}
	\caption{Cumulative distribution function (CDF) of RMSE metric on the test set of the evaluation datasets in Sec.~5 of the main paper. A point $(x,y)$ on the curve indicates the method achieve a recall of $y$ with threshold $x$ on that dataset. The diamonds show where the CDF reaches 1, i.e. the maximum error across the entire test set for that method. If there is no diamond, it means the maximum error is beyond the $x$-axis limit.}
	\label{fig:cdf}
\end{figure}

Fig.~\ref{fig:cdf} contains the cumulative distribution function (CDF) curves of the RMSE metric for the methods tested in Sec.~5 of the main paper. To be specific, a point $(x,y)$ on the curve implies that fraction $y$ of the instances in the test set has RMSE less than $x$. The CDF curves show the complete error distribution which reveals more information than a single metric. In fact, recall@0.2 shown in the main paper is a single point on the CDF curve with $x=0.2$. 

A couple of observations can be drawn from the error distribution
\begin{itemize}
    \item Some local methods such as ICP \cite{Chen92} and HGMR \cite{eckart2018hgmr} are quite accurate on a fraction of instances (in particular, those with small transformations).
    \item Methods based on point-to-point (ICP \cite{Chen92}, DCP \cite{wang2019deep}, PRNet \cite{wang2019prnet}) and feature correspondences (FGR \cite{zhou2016fast}, PointNetLK \cite{aoki2019pointnetlk}) performs worse on noisy data, whereas methods based on probabilistic data association (HGMR \cite{eckart2018hgmr}, DeepGMR) are unaffected.
    \item Learning-based methods (PointNetLK \cite{aoki2019pointnetlk}, DCP \cite{wang2019deep}, PRNet \cite{wang2019prnet}) except DeepGMR perform significantly worse on data from unseen categories, which shows that the generalization to unseen data demonstrated in these works does not hold in the case of unrestricted rotation.
\end{itemize}

\begin{figure}[h]
    \centering
	\begin{subfigure}{0.32\linewidth}
	    \centering
        \includegraphics[width=\linewidth]{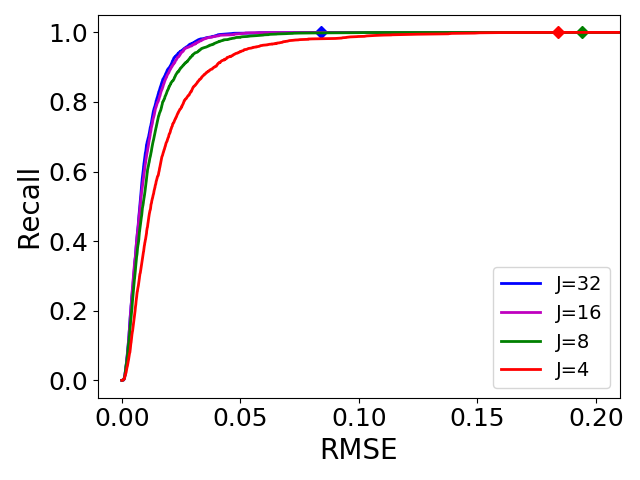}
        \vspace*{-7mm}
	    \caption{Number of components}
	\end{subfigure}
	\begin{subfigure}{0.32\linewidth}
	    \centering
        \includegraphics[width=\linewidth]{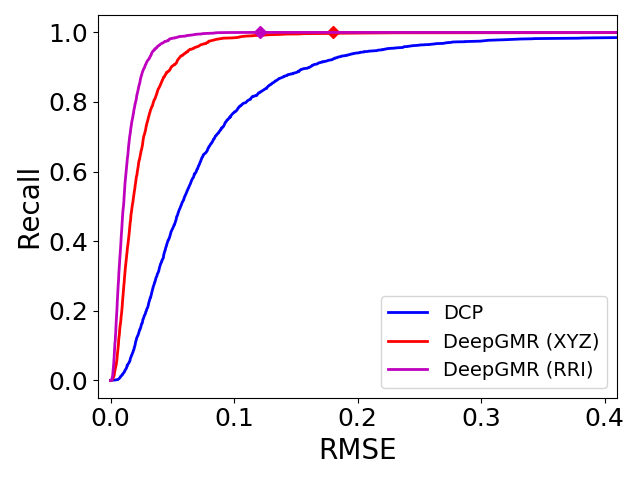}
        \vspace*{-7mm}
	    \caption{Input feature}
	\end{subfigure}
	\begin{subfigure}{0.32\linewidth}
	    \centering
        \includegraphics[width=\linewidth]{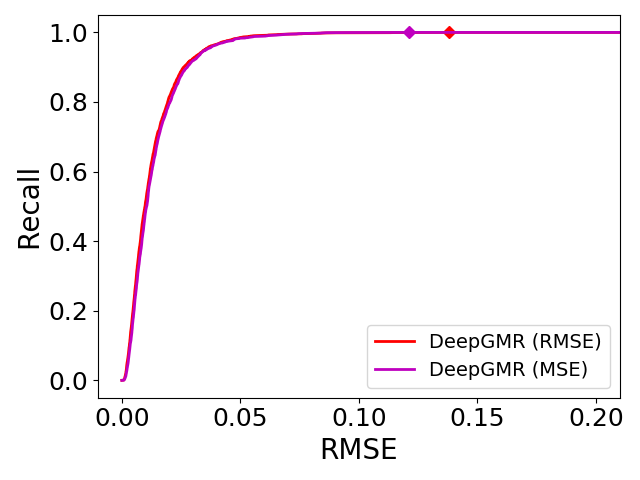}
        \vspace*{-7mm}
	    \caption{Loss function}
	\end{subfigure}
    \vspace*{-2mm}
	\caption{Ablation studies on \textbf{(a)} the number of GMM components $J$, \textbf{(b)} the input to correspondence network (RRI \cite{chen2019rri} or raw XYZ) and \textbf{(c)} loss function. Ablated models are compared using CDF of RMSE on the ModelNet noisy test set. The diamonds show where the CDF reaches 1, i.e. the maximum error across the entire test set for that method.}
	\label{fig:abl}
\end{figure}

\subsection{Ablation Studies}
We perform ablations on several design choices mentioned in Sec.~4 of the main paper, including the number of GMM components $J$, the input to correspondence network $f_\psi$ and the loss function. The dataset used is ModelNet noisy. The results are compared using CDF of RMSE on the test set shown in Fig.~\ref{fig:abl}.

\paragraph{Number of GMM components}
It can be seen that the performance of DeepGMR saturates with $J>16$, so we use $J=16$ across our experiments.

\paragraph{Input feature}
While the performance of DeepGMR is indeed improved with RRI features \cite{chen2019rri}, DeepGMR taking raw xyz coordinates still outperforms the most competitive baseline DCP. 

\paragraph{Loss function}
We trained DeepGMR directly with the RMSE metric used for evaluation (Eq.~20 of the main paper) and compared it to DeepGMR trained with the simple MSE loss in Eq.~18 of the main paper. We found the two models perform almost identically (the MSE-trained model has slightly lower maximum error). This shows that DeepGMR is not sensitive to the particular choice of loss function. Although it is possible to design a better loss function, it is not the focus of our work.

\subsection{Robustness Tests}
\begin{figure}[h]
    \centering
	\begin{subfigure}{0.45\linewidth}
	    \centering
        \includegraphics[width=\linewidth]{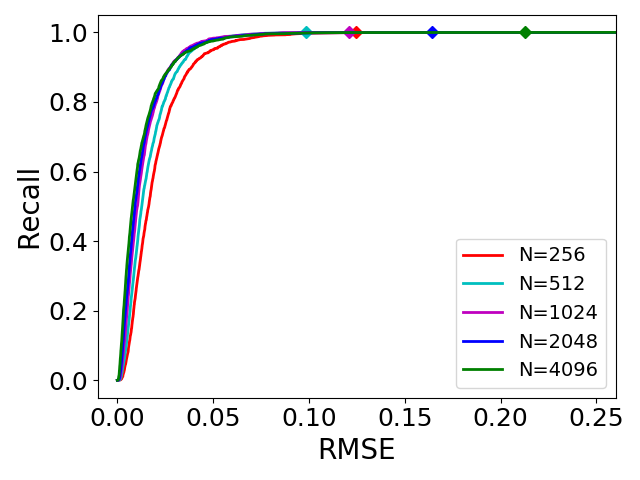}
        \vspace*{-7mm}
	    \caption{Point density}
	    \label{fig:density}
	\end{subfigure}
	\begin{subfigure}{0.45\linewidth}
	    \centering
        \includegraphics[width=\linewidth]{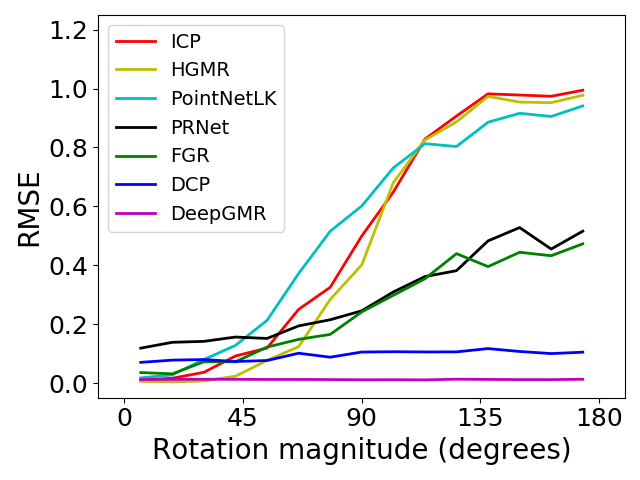}
        \vspace*{-7mm}
	    \caption{Transformation magnitude}
	    \label{fig:mag}
	\end{subfigure}
    \vspace*{-2mm}
	\caption{Robustness tests. \textbf{(a)} RMSE CDF curves of the same DeepGMR model (trained with $N=1024$) tested on point clouds of different density (density is measured by the number of subsampled points $N$). \textbf{(b)} RMSE vs magnitude of ground truth rotation between source and target on ModelNet noisy test set.}
	\label{fig:robust}
    \vspace*{-5mm}
\end{figure}

We perform additional tests on the robustness of DeepGMR to input point density and transformation magnitude. The results are shown in Fig.~\ref{fig:robust}.

\paragraph{Point density}
Because DeepGMR performs registration in the latent GMM space, it is invariant to the density of input point clouds. To demonstrate this, we test the DeepGMR model trained on ModelNet noisy in Sec.~5 of the main paper on point clouds with various density without any finetuning. Here, we can use the number of points $N$ as a proxy for density since the point clouds are uniformly sampled from same surface. From the results in Fig.~\ref{fig:density}, it can be seen that the performance of DeepGMR is unaffected on point clouds up to 4 times denser than training and is only slightly worse on point clouds up to 4 times sparser, which can be attributed to missing geometric details in sparse point clouds.

We note that the accuracy of methods that depend on hand-crafted feature correspondences, e.g. FGR, may improve with more input points as better normals can be estimated. However, we found that this is only true on data without noise. With 4096 input points, the accuracy of FGR improves on ModelNet clean (0.04 RMSE, 0.96 recall), but stays the same on ModelNet noisy (0.22 RMSE, 0.77 recall). This test indicates that FGR’s normal estimation accuracy is more contingent on the noise level than on the sampling density. 

\paragraph{Input transformation magnitude}
DeepGMR learns latent correspondences between points and GMM components that are \emph{pose-invariant}, which means that its output is invariant to the magnitude of the transformation between the input point clouds. From Fig.~\ref{fig:mag}, we can see that DCP \cite{wang2019deep}, another learning-based global method, shares the same invariance property while the performance of local methods (ICP \cite{Chen92}, HGMR \cite{eckart2018hgmr}, PointNetLK \cite{aoki2019pointnetlk}) degrades significantly with larger transformation. We also observe an interesting class of registration methods including FGR \cite{zhou2016fast} and PRNet \cite{wang2019prnet}. The formulations of these methods are global but they rely on feature matching or keypoint detection that become unstable with larger transformation, which makes their performance worse on these cases.

\subsection{Per-category Results}
Table~\ref{tab:cat} compares the average RMSE within each of the 40 categories in ModelNet noisy. DeepGMR achieves consistently good performance across all categories, while the baseline methods struggle with objects that have rotation or reflection symmetry (e.g. bowl, glass box), repetitive structure (e.g. bookshelf, stairs) or thin parts (e.g. radio, lamp).

%% file: supp_qual.tex
\section{Additional Qualitative Results} \label{supp:qual}
\subsection{Registration Results}
More qualitative registration results on ModelNet noisy and ICL-NUIM are shown in Fig.~\ref{fig:qual}. It can be seen that DeepGMR is able to deal with challenging cases that trap other methods in local minima, such as repetitive structure, undersampled thin parts and non-planar geometry, which demonstrate the consistency and robustness of the correspondences learned by DeepGMR.

\subsection{GMM Visualization}
We show more visualization of the learned GMM and correspondence in Fig.~\ref{fig:gmm}. We can see that different object parts are assigned to different GMM components consistently across views. Note that no explicit supervision is provided on the correspondence. Everything is learned end-to-end with the registration objective.

%% file: supp_future.tex
\section{Future Work} \label{supp:future}
One limitation of DeepGMR is that it does not explicitly consider partial overlap, i.e. when the IoU between source and target point clouds is less than 1 after alignment. The reason is that DeepGMR estimates the correspondence between \emph{all} points and \emph{all} components in the latent GMM. In the case of partial overlap, however, it is more ideal to estimate a partial correspondence, i.e. the correspondence between \emph{some} of the points and \emph{some} of the components in the latent GMM.

To measure the consequence of this limitation, we performed a preliminary experiment on partial data artificially created from ModelNet40. Specifically, we generate partial point clouds by approximating the rendering procedure of an orthographic depth camera. First, we randomly rotate the complete point cloud and project the points onto a zero-centered grid of dimension $200\times200$ and size $2\times2$ (same size as the bounding box of the point cloud, which is normalized to $[-1,1]^3$ across the dataset) on the $xy$-plane. Then, for each grid cell, we keep one point with the smallest $z$ value and throw away the others. In this way, we end up with a partial point cloud that closely resembles the observation of a depth camera. Finally, we add independent Gaussian noise to the points.

\begin{figure}[h]
    \centering
	\begin{subfigure}{0.48\linewidth}
	    \centering
	    \scriptsize
        \begin{tabular}{ccc}
            \toprule
            & RMSE $\downarrow$ & Re@0.2 $\uparrow$ \\ \midrule
            \red{ICP} \cite{Chen92} & 2.45 & 0.29 \\
            \red{HGMR} \cite{eckart2018hgmr} & 0.58 & 0.45 \\
            \red{PointNetLK} \cite{aoki2019pointnetlk} & 0.66 & 0.33 \\
            \red{PRNet} \cite{wang2019prnet} & 0.79 & 0.12 \\ \midrule
            \blue{FGR} \cite{zhou2016fast} & 0.50 & 0.43 \\
            \blue{DCP} \cite{wang2019deep} & 0.68 & 0.19 \\
            \blue{DeepGMR} & 0.46 & 0.34 \\ \midrule
            \orange{FGR+ICP} & 0.44 & 0.55 \\
            \orange{DCP+ICP} & 3.46 & 0.09 \\
            \orange{DeepGMR+ICP} & \textbf{0.34} & \textbf{0.64} \\
            \bottomrule
        \end{tabular}
	    \caption{RMSE and recall}
	\end{subfigure}
	\begin{subfigure}{0.46\linewidth}
	    \centering
        \includegraphics[width=\linewidth]{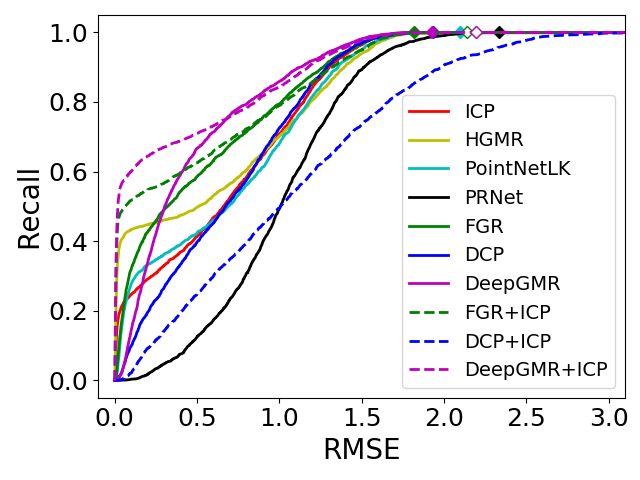}
        \vspace*{-7mm}
	    \caption{CDF curves}
	\end{subfigure}
    \vspace*{-3mm}
	\caption{Results on ModelNet partial: \textbf{(a)} Average RMSE and recall with threshold 0.2; \textbf{(b)} CDF of RMSE. \red{Local} methods outperform \blue{global} methods on a fraction of instances with small transformations but fail on the remaining ones. DeepGMR+ICP, a \orange{global+local} method that uses the output of DeepGMR as the initialization for ICP, achieves the best overall performance. Although DeepGMR by itself is not as accurate as in the case of complete overlap, it is able to bring most instances in the convergence basin of local methods. Best viewed in color.}
	\label{fig:partial}
\end{figure}

Experimental results on this partial dataset are summarized in Fig.~\ref{fig:partial}. On one hand, we note that even though DeepGMR does not explicitly consider partial overlap, its performance is still competitive. In addition, if we apply a refinement stage afterwards (i.e. use the prediction of a global method as the initialization of a local method such as ICP), DeepGMR achieves the best performance on this partial dataset. This demonstrates the power of the robust data association learned by DeepGMR. On the other hand, PRNet \cite{wang2019prnet}, a prior work that explicitly considers partial overlap, fails on this dataset. This shows that their method of dealing with partial overlap only works with limited transformation magnitude, i.e. it is a local registration method.

Although DeepGMR is able to outperform baselines on partial overlap data with the help of local refinement, its performance is still far below its performance on completely overlapping data. Therefore, a promising future research direction is to combine the robust point-to-latent-GMM correspondence learned by DeepGMR with techniques that deal with partial overlap (e.g. the attention mechanisms in \cite{wang2019deep,wang2019prnet}).

%% file: supp_fullpage.tex
\begin{figure*}
    \centering
    \tabcolsep 0pt
    \scriptsize
    \begin{tabular}{cccccccc}
        Input & ICP\cite{Chen92} & HGMR\cite{eckart2018hgmr} & PointNetLK \cite{aoki2019pointnetlk} & PRNet\cite{wang2019prnet} & FGR\cite{zhou2016fast} & DCP\cite{wang2019deep} & DeepGMR \\
        \includegraphics[width=0.12\linewidth]{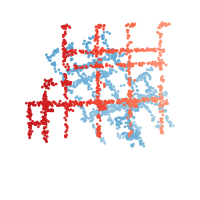} & \includegraphics[width=0.12\linewidth]{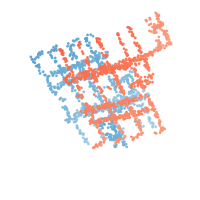} & \includegraphics[width=0.12\linewidth]{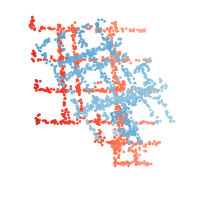} & \includegraphics[width=0.12\linewidth]{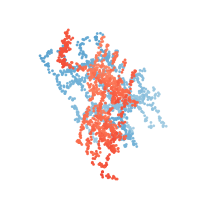} & \includegraphics[width=0.12\linewidth]{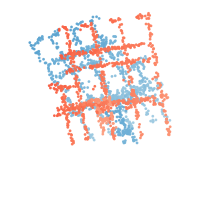} & \includegraphics[width=0.12\linewidth]{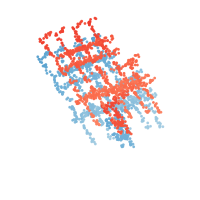} & \includegraphics[width=0.12\linewidth]{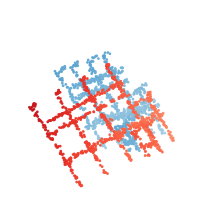} & \includegraphics[width=0.12\linewidth]{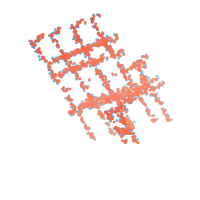} \\
        1.031 & 0.814 & 0.615 & 1.153 & 0.991 & 0.500 & 0.793 & 0.006 \\
        \includegraphics[width=0.12\linewidth]{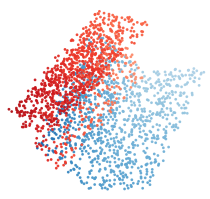} & \includegraphics[width=0.12\linewidth]{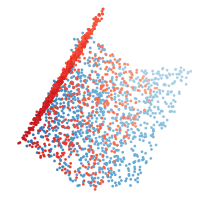} & \includegraphics[width=0.12\linewidth]{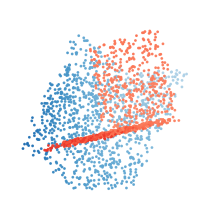} & \includegraphics[width=0.12\linewidth]{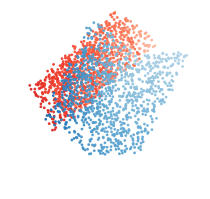} & \includegraphics[width=0.12\linewidth]{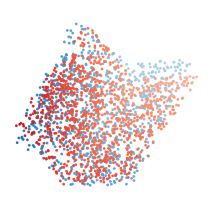} & \includegraphics[width=0.12\linewidth]{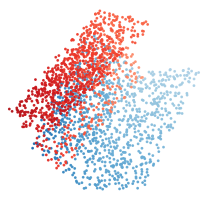} & \includegraphics[width=0.12\linewidth]{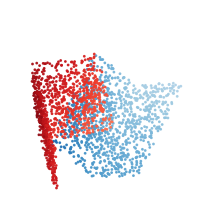} & \includegraphics[width=0.12\linewidth]{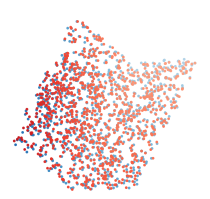} \\
        0.847 & 0.603 & 0.809 & 0.916 & 0.103 & 0.846 & 1.153 & 0.011 \\
        \includegraphics[width=0.12\linewidth]{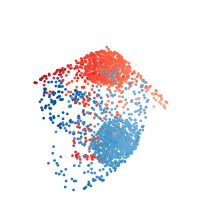} & \includegraphics[width=0.12\linewidth]{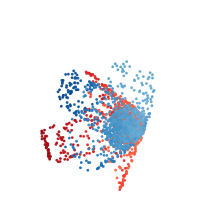} & \includegraphics[width=0.12\linewidth]{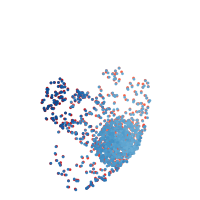} & \includegraphics[width=0.12\linewidth]{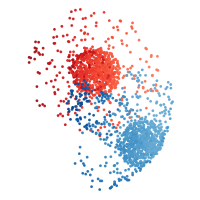} & \includegraphics[width=0.12\linewidth]{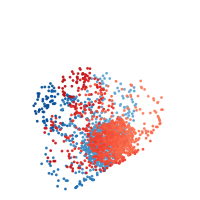} & \includegraphics[width=0.12\linewidth]{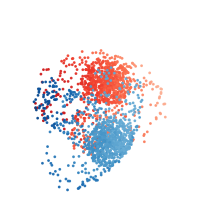} & \includegraphics[width=0.12\linewidth]{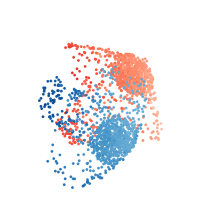} & \includegraphics[width=0.12\linewidth]{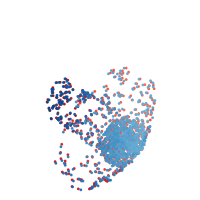} \\
        1.175 & 0.460 & 0.005 & 1.472 & 0.559 & 1.108 & 1.322 & 0.015 \\
        \includegraphics[width=0.12\linewidth]{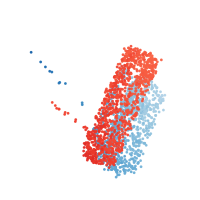} & \includegraphics[width=0.12\linewidth]{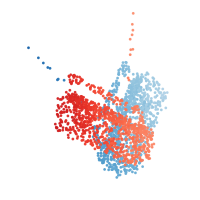} & \includegraphics[width=0.12\linewidth]{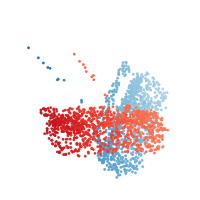} & \includegraphics[width=0.12\linewidth]{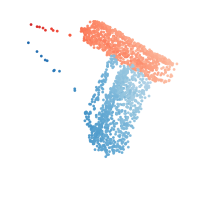} & \includegraphics[width=0.12\linewidth]{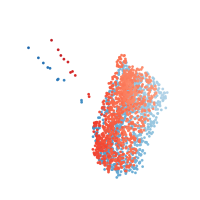} & \includegraphics[width=0.12\linewidth]{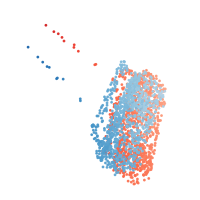} & \includegraphics[width=0.12\linewidth]{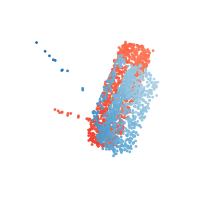} & \includegraphics[width=0.12\linewidth]{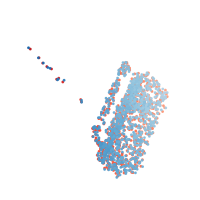} \\
        0.860 & 0.799 & 0.977 & 1.112 & 0.120 & 0.771 & 0.524 & 0.004 \\
        \includegraphics[width=0.12\linewidth]{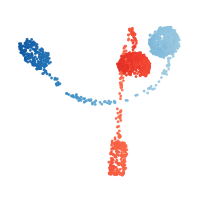} & \includegraphics[width=0.12\linewidth]{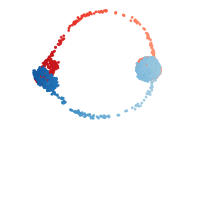} & \includegraphics[width=0.12\linewidth]{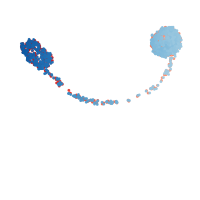} & \includegraphics[width=0.12\linewidth]{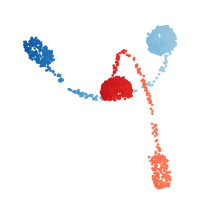} & \includegraphics[width=0.12\linewidth]{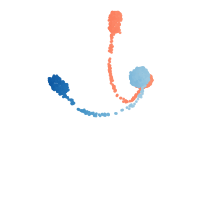} & \includegraphics[width=0.12\linewidth]{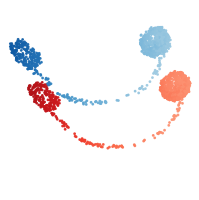} & \includegraphics[width=0.12\linewidth]{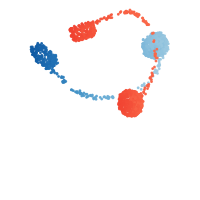} & \includegraphics[width=0.12\linewidth]{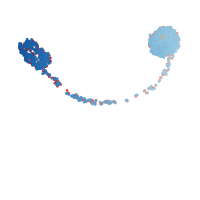} \\
        1.081 & 0.409 & 0.004 & 1.567 & 0.878 & 0.611 & 0.932 & 0.001 \\
        \includegraphics[width=0.12\linewidth]{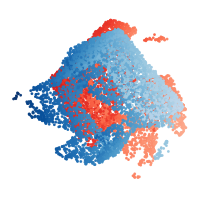} & \includegraphics[width=0.12\linewidth]{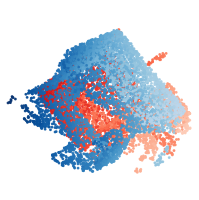} & \includegraphics[width=0.12\linewidth]{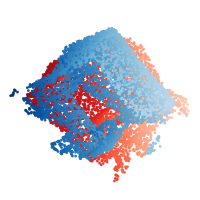} & \includegraphics[width=0.12\linewidth]{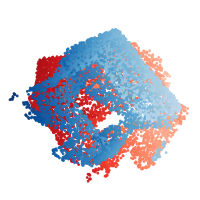} & \includegraphics[width=0.12\linewidth]{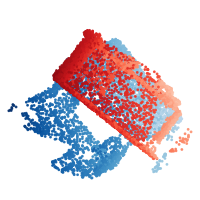} & \includegraphics[width=0.12\linewidth]{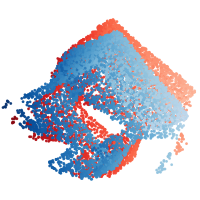} & \includegraphics[width=0.12\linewidth]{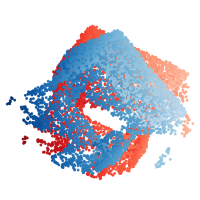} & \includegraphics[width=0.12\linewidth]{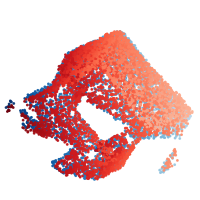} \\
        1.172 & 1.265 & 1.271 & 1.046 & 1.423 & 0.238 & 0.363 & 0.038 \\
        \includegraphics[width=0.12\linewidth]{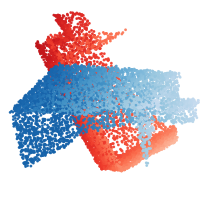} & \includegraphics[width=0.12\linewidth]{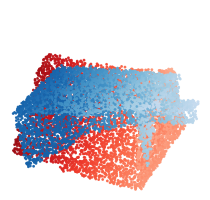} & \includegraphics[width=0.12\linewidth]{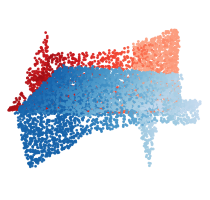} & \includegraphics[width=0.12\linewidth]{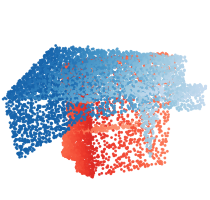} & \includegraphics[width=0.12\linewidth]{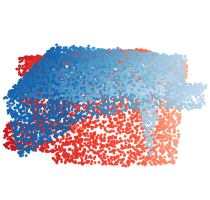} & \includegraphics[width=0.12\linewidth]{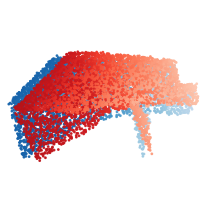} & \includegraphics[width=0.12\linewidth]{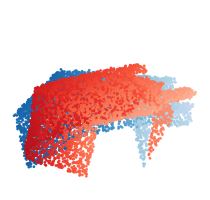} & \includegraphics[width=0.12\linewidth]{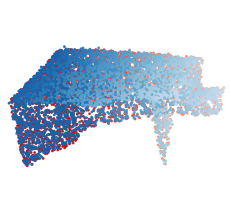} \\
        1.158 & 1.431 & 1.380 & 1.049 & 1.322 & 0.217 & 0.872 & 0.006
    \end{tabular}
    \caption{Qualitative registration results on ModelNet40 noisy (top 5 rows) and ICL-NUIM point clouds (bottom 2 rows). The RMSE of each example is labeled below the plot. These examples highlight some typical failure modes of existing methods such as 1) ignoring parts with sparse point samples 2) erroneous data association due to repetitive structures and symmetry. DeepGMR avoids these errors by estimating consistent point-to-distribution correspondence and performing robust registration on GMMs.}
    \label{fig:qual}
    \vspace{-.4cm}
\end{figure*}

\begin{figure}
    \centering
    \tabcolsep 5pt
    \begin{tabular}{cccc}
        Input & Correspondence & GMM & Output \\
        \multirow{2}{*}[7mm]{\includegraphics[width=0.22\linewidth]{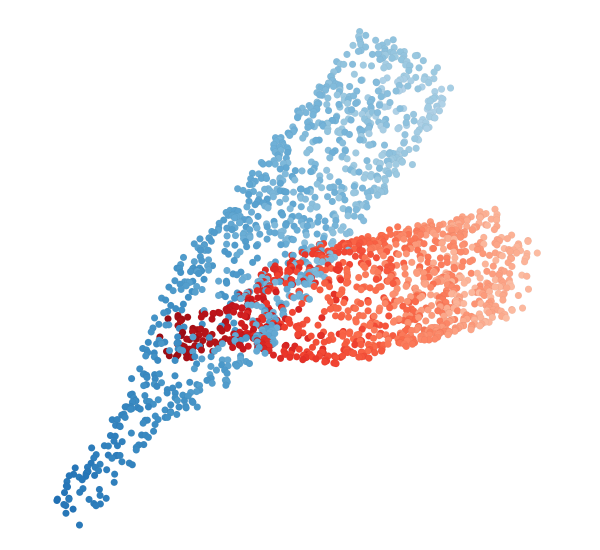}} & \includegraphics[width=0.16\linewidth]{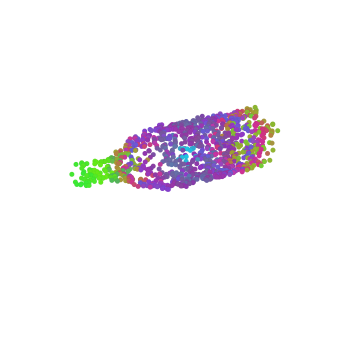} & \includegraphics[width=0.16\linewidth]{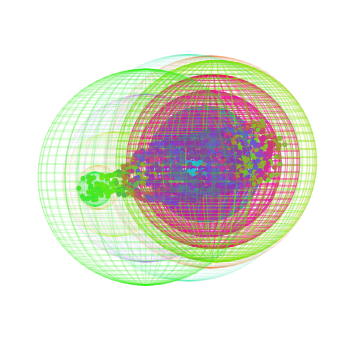} & \multirow{2}{*}[7mm]{\includegraphics[width=0.22\linewidth]{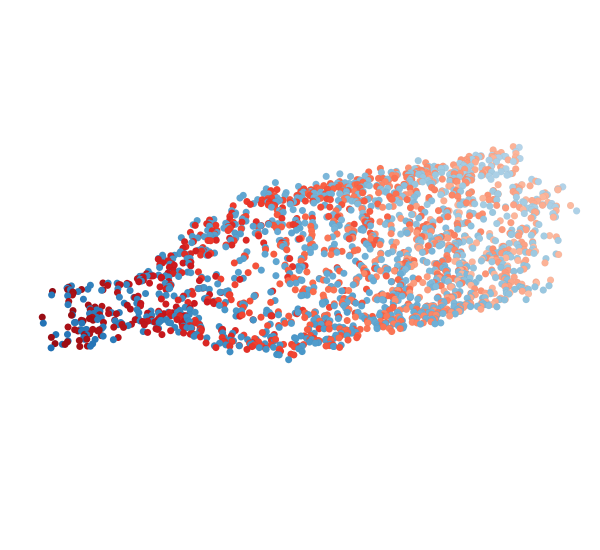}} \\
        & \includegraphics[width=0.16\linewidth]{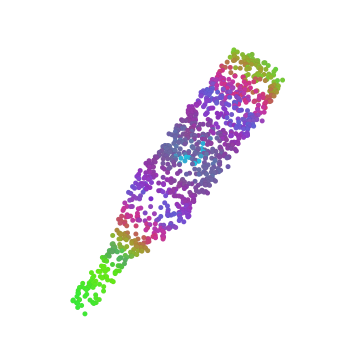} & \includegraphics[width=0.16\linewidth]{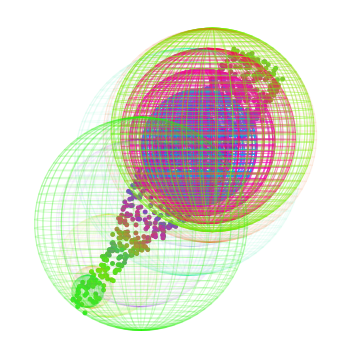} & \\
        \multirow{2}{*}[7mm]{\includegraphics[width=0.22\linewidth]{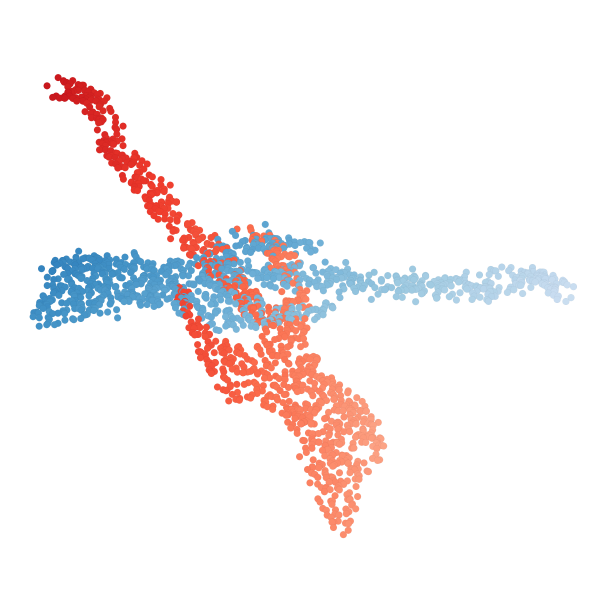}} & \includegraphics[width=0.16\linewidth]{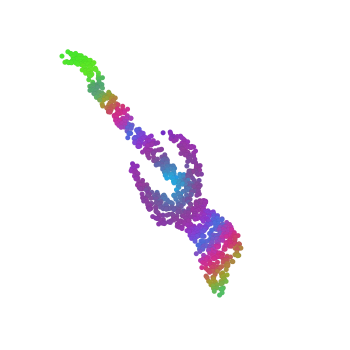} & \includegraphics[width=0.16\linewidth]{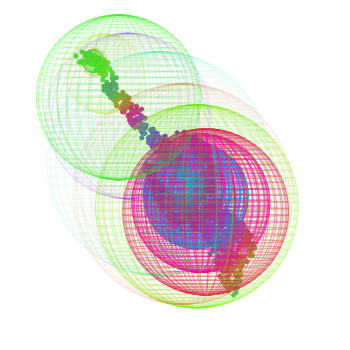} & \multirow{2}{*}[7mm]{\includegraphics[width=0.22\linewidth]{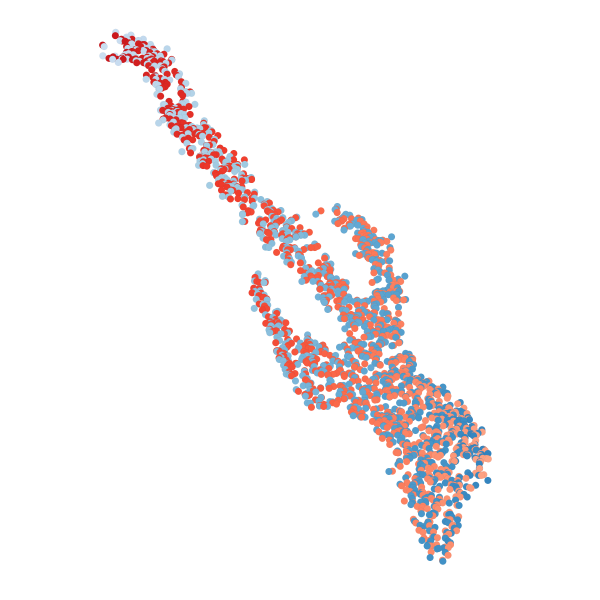}} \\
        & \includegraphics[width=0.16\linewidth]{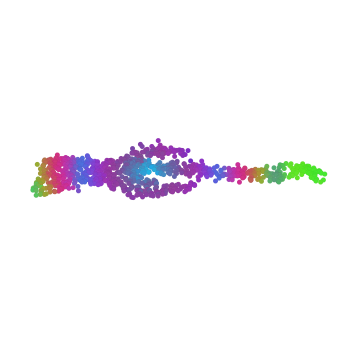} & \includegraphics[width=0.16\linewidth]{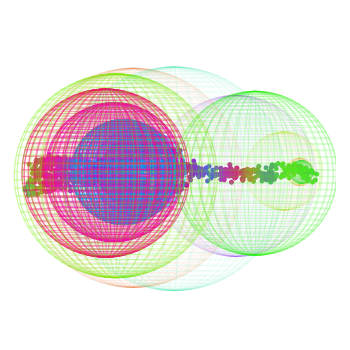} & \\
        \multirow{2}{*}[7mm]{\includegraphics[width=0.22\linewidth]{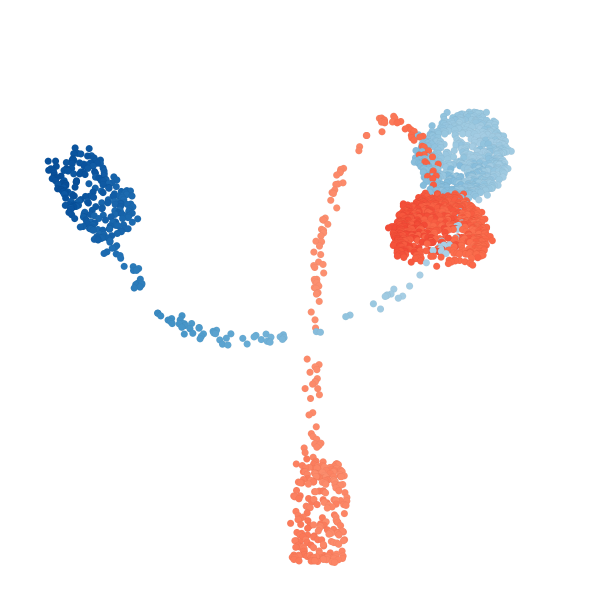}} & \includegraphics[width=0.16\linewidth]{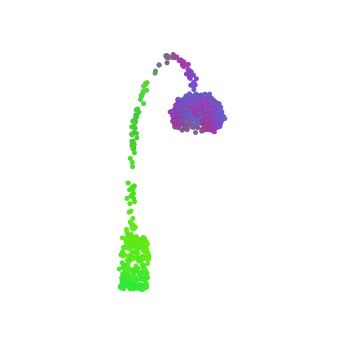} & \includegraphics[width=0.16\linewidth]{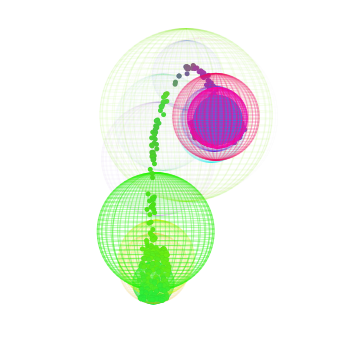} & \multirow{2}{*}[7mm]{\includegraphics[width=0.22\linewidth]{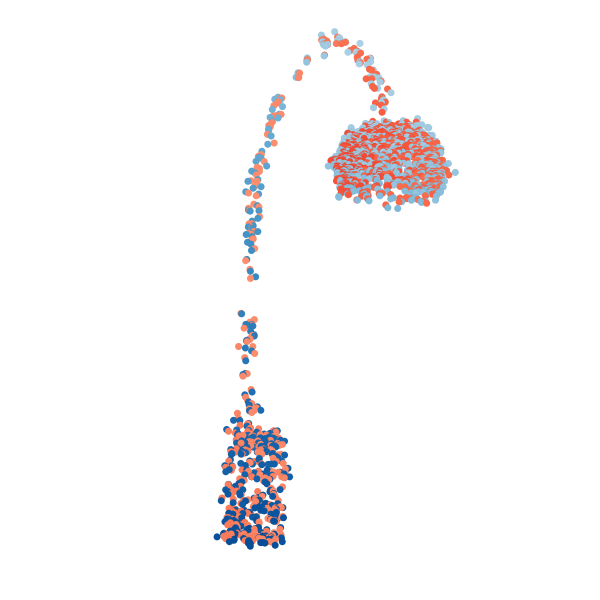}} \\
        & \includegraphics[width=0.16\linewidth]{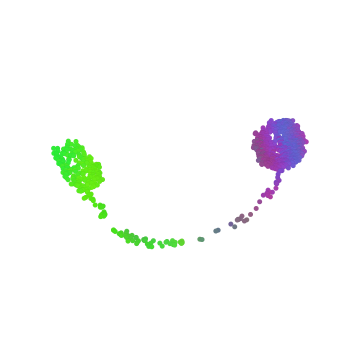} & \includegraphics[width=0.16\linewidth]{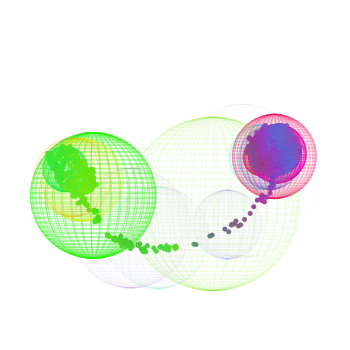} & \\
        \multirow{2}{*}[7mm]{\includegraphics[width=0.22\linewidth]{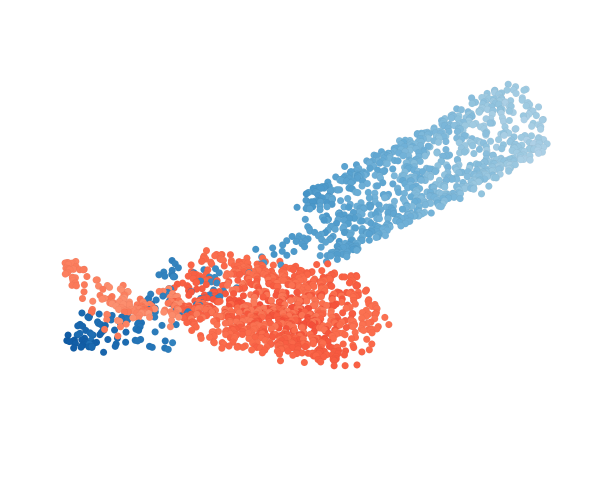}} & \includegraphics[width=0.16\linewidth]{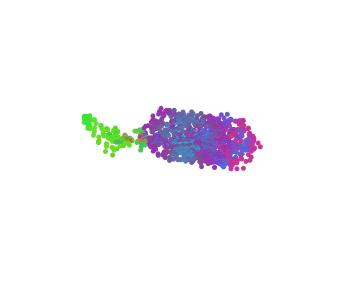} & \includegraphics[width=0.16\linewidth]{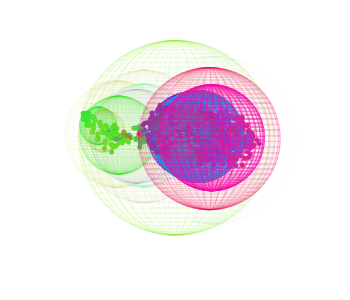} & \multirow{2}{*}[7mm]{\includegraphics[width=0.22\linewidth]{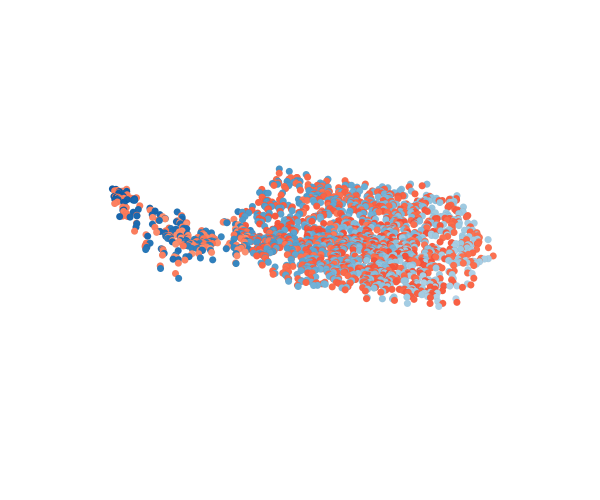}} \\
        & \includegraphics[width=0.16\linewidth]{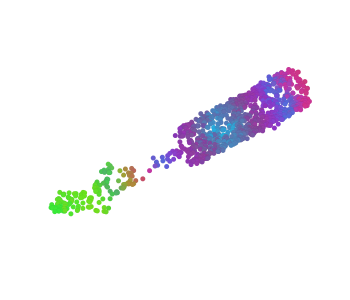} & \includegraphics[width=0.16\linewidth]{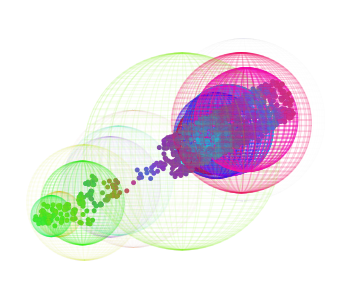} &
    \end{tabular}
    \vspace{-5mm}
	\caption{Visualization of learned correspondences and GMMs. In the second and third column, each color indicates a different GMM component. The point colors are calculated as weighted averages of the component colors according to the learned correspondences. The radius of each sphere is equal to the standard deviation of the GMM component. Note how DeepGMR learns to correspond the points and GMMs in the source (red in first column, top in each row) and target (blue in first column, bottom in each row) without any explicit supervision.}
	\label{fig:gmm}
\end{figure}

\begin{table}[h]
    \centering
    \caption{Comparison of average RMSE within each category on ModelNet noisy}
    \label{tab:cat}
    \vspace{2mm}
    \begin{tabular}{cccccccc}
        Category & ICP\cite{Chen92} & HGMR\cite{eckart2018hgmr} & PointNetLK \cite{aoki2019pointnetlk} & PRNet\cite{wang2019prnet} & FGR\cite{zhou2016fast} & DCP\cite{wang2019deep} & DeepGMR \\ \midrule
        airplane & 0.49 & 0.48 & 0.51 & 0.24 & 0.15 & 0.08 & \textbf{0.01} \\
        bathtub & 0.53 & 0.43 & 0.62 & 0.31 & 0.30 & 0.13 & \textbf{0.01} \\
        bed & 0.46 & 0.66 & 0.52 & 0.20 & 0.24 & 0.08 & \textbf{0.01} \\
        bench & 0.51 & 0.62 & 0.42 & 0.41 & 0.38 & 0.15 & \textbf{0.02} \\
        bookshelf & 0.58 & 0.50 & 0.56 & 0.33 & 0.22 & 0.14 & \textbf{0.01} \\
        bottle & 0.51 & 0.55 & 0.51 & 0.27 & 0.25 & 0.07 & \textbf{0.01} \\
        bowl & 0.77 & 0.76 & 0.80 & 0.57 & 0.61 & 0.15 & \textbf{0.02} \\
        car & 0.41 & 0.44 & 0.47 & 0.23 & 0.16 & 0.09 & \textbf{0.01} \\
        chair & 0.55 & 0.51 & 0.55 & 0.25 & 0.20 & 0.09 & \textbf{0.01} \\
        cone & 0.50 & 0.77 & 0.54 & 0.24 & 0.30 & 0.13 & \textbf{0.02} \\
        cup & 0.64 & 0.67 & 0.72 & 0.48 & 0.39 & 0.10 & \textbf{0.01} \\
        curtain & 0.58 & 0.41 & 0.41 & 0.34 & 0.36 & 0.10 & \textbf{0.01} \\
        desk & 0.58 & 0.54 & 0.53 & 0.21 & 0.18 & 0.11 & \textbf{0.01} \\
        door & 0.55 & 0.53 & 0.55 & 0.26 & 0.60 & 0.14 & \textbf{0.02} \\
        dresser & 0.57 & 0.51 & 0.59 & 0.34 & 0.23 & 0.10 & \textbf{0.02} \\
        flower pot & 0.27 & 0.51 & 0.37 & 0.25 & 0.15 & 0.09 & \textbf{0.01} \\
        glass box & 0.65 & 0.68 & 0.61 & 0.41 & 0.47 & 0.11 & \textbf{0.02} \\
        guitar & 0.40 & 0.52 & 0.36 & 0.31 & 0.47 & 0.05 & \textbf{0.00} \\
        keyboard & 0.55 & 0.44 & 0.53 & 0.47 & 0.45 & 0.07 & \textbf{0.01} \\
        lamp & 0.59 & 0.78 & 0.37 & 0.27 & 0.18 & 0.07 & \textbf{0.01} \\
        laptop & 0.32 & 0.53 & 0.51 & 0.37 & 0.37 & 0.09 & \textbf{0.01} \\
        mantel & 0.56 & 0.48 & 0.56 & 0.27 & 0.23 & 0.08 & \textbf{0.01} \\
        monitor & 0.61 & 0.48 & 0.59 & 0.27 & 0.30 & 0.10 & \textbf{0.01} \\
        night stand & 0.63 & 0.58 & 0.56 & 0.36 & 0.27 & 0.11 & \textbf{0.01} \\
        person & 0.43 & 0.27 & 0.42 & 0.27 & 0.19 & 0.07 & \textbf{0.00} \\
        piano & 0.55 & 0.58 & 0.59 & 0.24 & 0.15 & 0.08 & \textbf{0.01} \\
        plant & 0.51 & 0.50 & 0.47 & 0.27 & 0.21 & 0.07 & \textbf{0.01} \\
        radio & 0.37 & 0.71 & 0.43 & 0.26 & 0.33 & 0.09 & \textbf{0.01} \\
        range hood & 0.56 & 0.50 & 0.54 & 0.31 & 0.25 & 0.07 & \textbf{0.01} \\
        sink & 0.51 & 0.60 & 0.54 & 0.30 & 0.16 & 0.09 & \textbf{0.01} \\
        sofa & 0.43 & 0.50 & 0.49 & 0.35 & 0.18 & 0.10 & \textbf{0.01} \\
        stairs & 0.63 & 0.53 & 0.58 & 0.29 & 0.46 & 0.15 & \textbf{0.01} \\
        stool & 0.76 & 0.71 & 0.63 & 0.20 & 0.25 & 0.13 & \textbf{0.01} \\
        table & 0.59 & 0.46 & 0.60 & 0.34 & 0.51 & 0.14 & \textbf{0.01} \\
        tent & 0.52 & 0.53 & 0.51 & 0.29 & 0.21 & 0.10 & \textbf{0.01} \\
        toilet & 0.53 & 0.49 & 0.48 & 0.23 & 0.11 & 0.08 & \textbf{0.01} \\
        tv stand & 0.47 & 0.50 & 0.50 & 0.29 & 0.20 & 0.09 & \textbf{0.01} \\
        vase & 0.61 & 0.65 & 0.64 & 0.36 & 0.34 & 0.12 & \textbf{0.02} \\
        wardrobe & 0.65 & 0.63 & 0.61 & 0.36 & 0.43 & 0.10 & \textbf{0.02} \\
        xbox & 0.46 & 0.57 & 0.47 & 0.28 & 0.25 & 0.06 & \textbf{0.01} \\
    \end{tabular}
\end{table}